%% file: main.tex
\documentclass{article}
\usepackage{amssymb}

\usepackage[preprint]{corl_2025} % Uncomment for pre-prints (e.g., arxiv); This is like ``final'', but will remove the CORL footnote.

\usepackage{booktabs}
\usepackage{comment}

\usepackage[utf8]{inputenc} % allow utf-8 input
\usepackage[T1]{fontenc}    % use 8-bit T1 fonts
\usepackage{url}            % simple URL typesetting
\usepackage{booktabs}
\usepackage{multirow}
\usepackage{graphicx}
\usepackage{amsfonts}       % blackboard math symbols
\usepackage{nicefrac}       % compact symbols for 1/2, etc.
\usepackage{microtype}      % microtypography
\usepackage{tabularx}

\usepackage{microtype}
\usepackage{graphicx}
\usepackage{subcaption}
\usepackage{hyperref}
\hypersetup{
	colorlinks=true,
	linkcolor=orange,
	filecolor=magenta,      
	urlcolor=orange,
	citecolor=orange,
}
\usepackage[utf8]{inputenc} % allow utf-8 input
\usepackage[T1]{fontenc}    % use 8-bit T1 fonts
\usepackage{dsfont}
\usepackage{url}            % simple URL typesetting
\usepackage{amsfonts}       % blackboard math symbols
\usepackage{nicefrac}       % compact symbols for 1/2, etc.
\usepackage{color}
\usepackage[dvipsnames,table]{xcolor}
\usepackage{mathtools}
\usepackage{amsmath,amssymb}
\usepackage{bm}
\usepackage{siunitx}
\usepackage{wrapfig}
\usepackage{algorithm}
\usepackage[noend]{algorithmic}
\usepackage{lipsum}
\usepackage{makecell}
\usepackage{cite}
\usepackage{subcaption}
\usepackage[capitalise, nameinlink]{cleveref}

\usepackage{footnote}
\makesavenoteenv{tabular}
\makesavenoteenv{table}
\usepackage{multirow}
\usepackage{booktabs}
\usepackage{amsmath}
\usepackage{amssymb}
\usepackage{mathtools}
\usepackage{listings}

\usepackage{pifont}% http://ctan.org/pkg/pifont

\hypersetup{
    colorlinks = true,
    citecolor = {magenta},
}

\definecolor{burntorange}{rgb}{0.8, 0.33, 0.0}
\definecolor{bluegray}{rgb}{0.4, 0.6, 0.8}
\definecolor{asparagus}{rgb}{0.53, 0.66, 0.42}

\newcommand\blfootnote[1]{%
  \begingroup
  \renewcommand\thefootnote{}\footnote{#1}%
  \addtocounter{footnote}{-1}%
  \endgroup
}

\usepackage[most]{tcolorbox}
\definecolor{myblue}{HTML}{598BE7}

\def \Acronym {ECoT-Lite}

\title{Training Strategies for Efficient Embodied Reasoning}

\author{
William Chen$^{1}$, Suneel Belkhale$^{2}$, Suvir Mirchandani$^{2}$ \\
\textbf{Oier Mees$^{1}$, Danny Driess$^{3}$, Karl Pertsch$^{1}$, Sergey Levine$^{1}$} \\
${}^{1}$UC Berkeley, ${}^{2}$Stanford University, ${}^{3}$Physical Intelligence \\
\href{https://ecot-lite.github.io}{ecot-lite.github.io}
}

\begin{document}
\maketitle

\begin{abstract}
    Robot chain-of-thought reasoning (CoT) -- wherein a model predicts helpful intermediate representations before choosing actions -- provides an effective method for improving the generalization and performance of robot policies, especially vision-language-action models (VLAs). While such approaches have been shown to improve performance and generalization, they suffer from core limitations, like needing specialized robot reasoning data and slow inference speeds. To design new robot reasoning approaches that address these issues, a more complete characterization of why reasoning helps policy performance is critical. We hypothesize several mechanisms by which robot reasoning improves policies -- (1) better representation learning, (2) improved learning curricularization, and (3) increased expressivity -- then devise simple variants of robot CoT reasoning to isolate and test each one. We find that learning to generate reasonings does lead to better VLA representations, while attending to the reasonings aids in actually leveraging these features for improved action prediction. Our results provide us with a better understanding of why CoT reasoning helps VLAs, which we use to introduce two simple and lightweight alternative recipes for robot reasoning. Our proposed approaches achieve significant performance gains over non-reasoning policies, state-of-the-art results on the LIBERO-90 benchmark, and a 3x inference speedup compared to standard robot reasoning.
\end{abstract}


\blfootnote{Correspondence to: \href{mailto:verityw@berkeley.edu}{\texttt{verityw@berkeley.edu}}}

%===============================================================================
\vspace{-0.3cm}
\begin{figure}[H]
\begin{minipage}{\textwidth}\includegraphics[width=\linewidth]{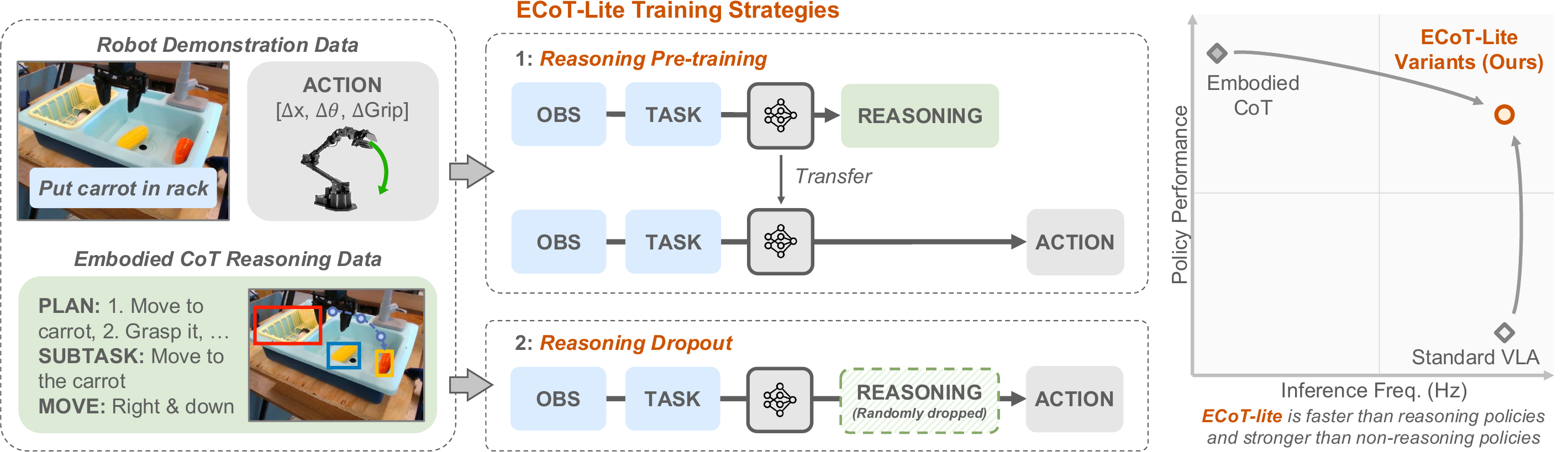}
\caption{\footnotesize{
        Illustration of our proposed \Acronym{} approaches. Past robot reasoning policies are performant but slow. By testing numerous hypotheses on why robot reasoning improves policy performance, we find two simple lightweight alternatives for training policies with embodied reasoning data \textit{without} producing reasonings at test time, boosting performance over non-reasoning VLAs while maintaining fast inference speeds.}}
\label{fig:teaser}
\end{minipage}
\end{figure}
\vspace{-0.5cm}

\vspace{-0.3cm}
\section{Introduction}
\label{sec:intro}
\vspace{-0.3cm}

Useful robots must generalize to a broad range of real-world scenarios. As such, policy \emph{generalization} has been a long-standing focus for the robotics community. Recent works have shown that training policies on large and diverse datasets of robot experience can significantly improve their generalization ability \citep{EmbodimentCollaboration24-oxe, Brohan23-rt1, OMT23-octo,Doshi24-crossformer}. Vision-language-action models (VLAs \citep{Brohan23-rt2, Kim24-openVLA}) have been proposed as a particularly promising approach for generalizable policy training: by fine-tuning a pre-trained vision-language model (VLM) with diverse robot control data, these models combine the scalability of large transformer architectures with rich semantic knowledge from the VLM's internet pre-training. 

While the standard approach to improving policy generalization in such a data-driven regime is to collect increasingly large robot datasets, often through tedious human teleoperation \citep{EmbodimentCollaboration24-oxe, Ebert21-bridgeDataV1, Walke23-bridgeDataV2,rosete2022corl}, the flexibility of the VLA architecture~\citep{Black24-pi0,bjorck2025gr00t,jones24fuse,zhen20243d} has recently given rise to an alternative, \emph{orthogonal} paradigm for improving generalization: embodied chain-of-thought reasoning (ECoT \citep{Zawalski24-ecot}). Inspired by work in reasoning for large language models \citep{Wei23-chainOfThought, Kojima23-zeroShotChainOfThought}, these approaches train VLAs to split the robot action prediction problem into a sequence of reasoning steps, like identifying the locations of objects and the robot's end-effector~\citep{zheng2024tracevla}, planning subtasks \citep{Brohan23-rt2}, or predicting object affordances~\citep{xu2025a0,borja22affordance}. Importantly, there is ample evidence that training robot policies to perform such reasoning steps can significantly improve their ability to generalize to new scenes, objects, and task instructions \citep{Zawalski24-ecot}, \emph{without} the need to collect additional robot demonstrations.

However, while embodied chain-of-thought reasoning is a promising approach for improving policy generalization, existing robot reasoning approaches come with significant costs: training data needs to be annotated with detailed reasoning instructions, and performing extended reasoning steps during inference can significantly slow down policy rollouts, to the point where a single action prediction may take multiple seconds \citep{Zawalski24-ecot}. The latter downside has made it particularly challenging to use sophisticated robot reasoning approaches in practice.

In this work, we develop more practical strategies for training policies with embodied reasoning data. Our key observation is that there are multiple distinct hypotheses for \emph{why} embodied chain-of-thought reasoning may help generalization, e.g., it may improve the policy's learned representations or it may increase the effective model capacity through increased inference compute. Based on this observation, we develop a number of novel, lightweight embodied chain-of-thought recipes, ``\Acronym{}'', that isolate the effects of these hypotheses. Crucially, these recipes retain most of the generalization benefits of regular chain-of-thought policy training while avoiding its drawbacks, making our \Acronym{} approaches significantly more practical. Concretely, \Acronym{} achieves state-of-the-art performance on the widely used LIBERO simulation benchmark \citep{Liu23-libero}, and outperforms state-of-the-art conventional VLAs on BridgeData~V2 evaluations by 10-19\%, while speeding up inference from 1-1.2 Hz for conventional embodied chain-of-thought reasoning approaches to the 3.5+ Hz of standard VLAs. In doing so, we also develop a deeper understanding of why embodied chain-of-thought works, which we hope will inspire future robot reasoning research.

In summary, the core contributions of our work are:
\textbf{(1)} a detailed empirical analysis of the mechanisms by which embodied chain-of-thought reasoning improves policy performance, from which we develop \textbf{(2)} a set of novel, simplified strategies for training robot reasoning policies, which we validate with \textbf{(3)} comprehensive simulated and real-robot evaluations that show that our simplified robot reasoning recipes retain substantially improved generalization performance while making robot reasoning policies significantly more practical to use.

\vspace{-0.3cm}
\section{Related Work}
\label{sec:related_work}
\vspace{-0.4cm}

\textbf{Vision-language-action models.}
Vision language action models (VLAs) are a way to train robot policies by adapting vision-language models (VLMs) \citep{Karamcheti24-prismatic, Beyer24-paligemma, Steiner24-paligemma2, Dai23-instructblip} to output robot actions \citep{Kim24-openVLA, Brohan23-rt2, szot2024multimodal, Black24-pi0, huang2025otter, wen2024tinyvlafastdataefficientvisionlanguageaction, liu2025hybridvla, liu2024rdt, li2024cogact, Belkhale24-miniVLA, Pertsch25-fast, GRT25-geminirobotics, wen2025dexvla, zhen20243d, bjorck2025gr00t, chi2024universal, jones24fuse, etukuru2024robotutilitymodelsgeneral, shafiullah2023bringingrobotshome, lin2024data}.
Several works have shown that by framing action prediction as a vision-language problem, the model benefits from the internet-scale pre-training on general language and vision-language tasks, leading to better robustness and generalization capabilities \citep{Brohan23-rt2}.
VLAs are trained via behavioral cloning (BC) on large datasets \citep{EmbodimentCollaboration24-oxe, Ebert21-bridgeDataV1, Walke23-bridgeDataV2, Khazatsky24-droid}, typically representing actions as discrete tokens.

\textbf{Chain-of-thought in LLMs and VLMs.}
Many studies have shown that chain-of-thought reasoning (CoT) -- wherein a model generates intermediate reasoning steps before producing an answer to a query \citep{Wei23-chainOfThought, Kojima23-zeroShotChainOfThought} -- can significantly improve the performance of LLMs and VLMs.
The intuition is that each intermediate step may be easier for a model to predict compared to directly generating the correct answer.
CoT has been particularly successful for tasks involving mathematics \citep{sprague2024cot, Snell24-scalingTestTimeCompute}, where this intuition is theoretically justified: the marginal of a fix-sized transformer architecture is not expressive enough to solve problems like integer arithmetic or dynamic programming, but employing suitably-long CoT allows LLMs to solve these problems \citep{Feng23-revealingmysteryCoT}. Other analyses of why reasoning helps LLMs make similar arguments, showing that intermediate reasoning steps increase the expressivity of a transformer \citep{Merrill24-CoTExpressivePower, Li24-CoTSerialProblems}.
While these theoretical analyses give insights on the complexity of logical problems that can be solved with CoT, this view is incomplete, since it does not address why reasoning empirically helps with \emph{common-sense} problem solving (like VQA or robotics) \citep{Wei23-chainOfThought}, which differs from math problems in that it requires connecting fragmented or implicit knowledge about the real world instead of logical deduction over well-defined operations on abstract symbols.
In this work, we aim to fill this gap by investigating several hypotheses about why CoT benefits real-world robotics problems. We focus specifically on tasks that rely on semantic reasoning, similar to common-sense question answering, as distinct from reasoning over logical symbols.

 \begin{wrapfigure}{R}{0.4\textwidth}
    \centering
    \vspace{-0.5cm}
    \includegraphics[width=0.39\textwidth]{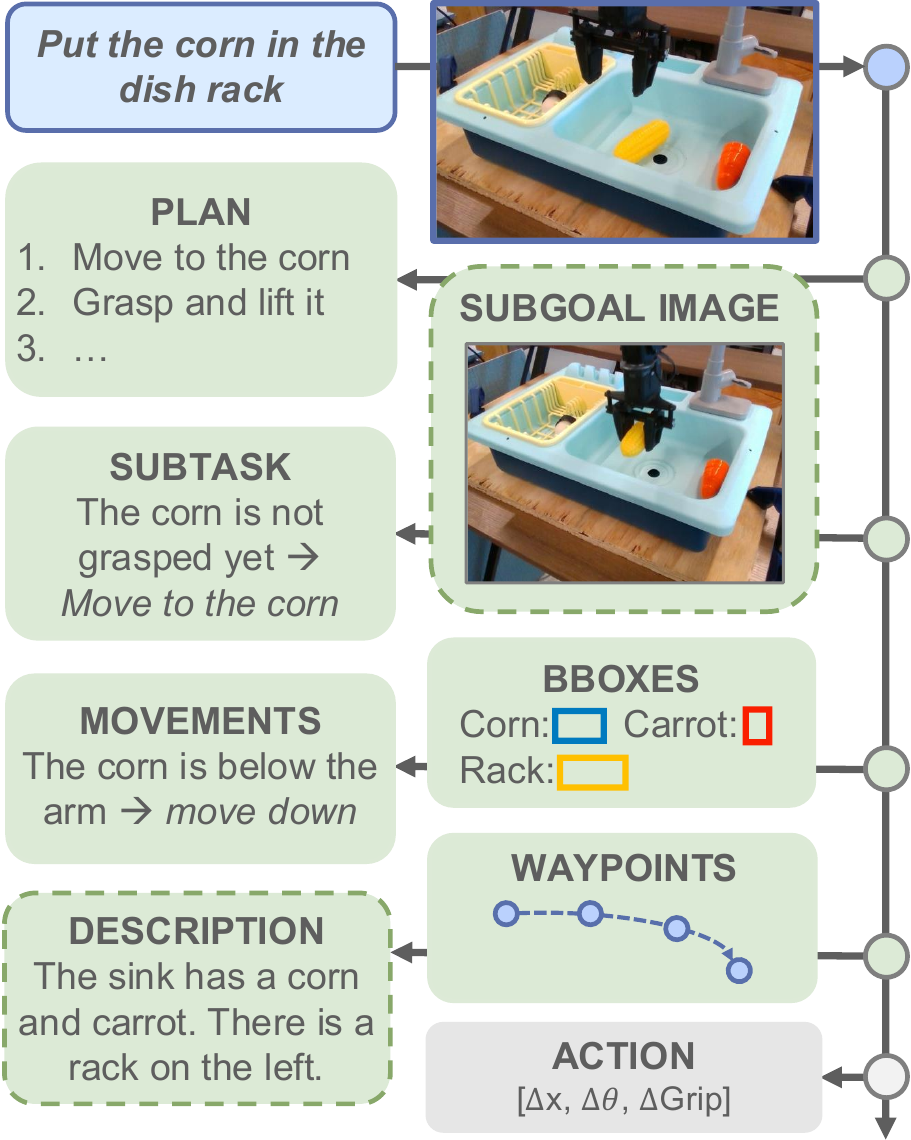}
    \vspace{-0.25cm}
    \caption{
        \footnotesize{
        Example intermediate reasoning steps. We use Embodied Chain-of-Thought Reasoning (ECoT \citep{Zawalski24-ecot}) as a representative robot reasoning approach for this work, and thus indicate which steps it does \textit{not} use with dashed borders (but they are used in other similar works \citep{Hwang24-emma, Zhao25-cotvla-visualchainofthoughtreasoning}).
        }
    }
    \vspace{-.5cm}
    \label{fig:robot-reasoning-schematic}
\end{wrapfigure}

\textbf{Robot reasoning.}
Inspired by the success of CoT for LLMs, there are several works that try to leverage CoT reasoning for robotics \citep{Zawalski24-ecot, Zhao25-cotvla-visualchainofthoughtreasoning, Clark25-actionFreeReasoning, Brohan23-rt2}.
Instead of mapping observations directly to robot actions as in standard VLAs, CoT in robotics first generates robot ``reasonings'' before predicting the final actions
\citep{Zawalski24-ecot}. Typical reasoning steps include breaking down goals into subtasks \citep{Sharma22-skillInductionLatentLanguage, Brohan23-rt2, Huang22-innerMonologue,liang2023code,mees23hulc2}; grounded features including representations of robot goals or motions \citep{Belkhale24-rth, Gu23-rttrajectory, Niu24-llarva, Zhao25-cotvla-visualchainofthoughtreasoning,langsurvery24ijrr}; and task-relevant visuo-semantic features \citep{Chen24-pr2l, Zawalski24-ecot} such as object bounding boxes or semantic keypoints.
Prior works show that predicting these intermediate reasonings improves policy generalization compared to training only with action prediction, but fail to dissect \emph{why} this is the case. Compared to LLMs, robot reasoning in VLAs cannot usually be elicited by prompting alone. Instead, the model is explicitly trained to produce reasonings \citep{Zawalski24-ecot}, which raises the question to what extent the CoT itself or the additional training signal contributes to the increased performance.
Our work investigates this by dissecting the different aspects of CoT both at training and inference time.

\textbf{Robot representation learning.}
Training a VLA to reason can be interpreted as a form of representation learning, which has a long history in robot learning.
Numerous authors have suggested pre-training base models on egocentric/embodied data \citep{Majumdar24-vc1, Karamcheti23-voltron} or grounded spatial reasoning tasks \citep{Chen24-spatialvlm, Driess23-palme}, as the resulting representations might be more conducive to learning control. \citet{Burns23-whatMakesPretrainedVisualRepresentations} provide experimental support that representations conducive to spatial-semantic understanding tasks (e.g., segmentation) are indicative of their visual robustness in robot policies. Further, a core motivation behind VLAs is that VLMs learn generalizable visuo-linguistic representations during pre-training \citep{Brohan23-rt2}. Even a frozen VLM's representations can be used to learn policies, with performance increasing further if representations of reasoning are included \citep{Chen24-pr2l}. However, when analyzing a wider set of generalization axes, \citet{gao2025taxonomy} found that co-training with VLM training data was not uniformly beneficial for all types of generalization.
Finally, \citet{GRT25-geminirobotics} connects embodied representation learning and reasoning by both training their base VLM on embodied data and then fine-tuning it into a reasoning robot policy.

In summary, previous work either focuses on representation learning, theoretical analysis of CoT for math problems, or simply shows that CoT improves performance for robotics.
To our knowledge, our work is the first to dissect why training on and using CoT at inference time is beneficial in embodied robotics settings, investigating both the representation learning and test-time compute aspects.

\vspace{-0.2cm}
\section{Preliminaries}
\label{sec:prelim}
\vspace{-0.5cm}

\textbf{Vision-language-action models.}
Given a task expressed in natural language $\ell$, obtaining a robot policy can be seen as learning a function $\pi$ from which we can sample actions $a \sim \pi(\cdot | \ell, o)$ conditioned on the robot observation $o$ (e.g.\ image observations or the robot's proprioceptive state).
The idea of vision-language-action models is to represent robot actions as a sequence of text tokens, for example through simple discretization and binning strategies, such that a policy can be trained by finetuning a vision-language model into outputting robot actions autoregressively.

\textbf{Embodied chain-of-thought reasoning.} We consider \textit{Embodied Chain-of-Thought Reasoning} (ECoT) as a representative robot reasoning approach for this work \citep{Zawalski24-ecot}. ECoT makes a simple change to the VLA formulation: the policy first generates reasoning text before predicting the action tokens. Possible reasoning steps include both high-level planning (such as subtask prediction) and low-level grounded features such as motions, gripper positions, and object bounding boxes (see \cref{fig:robot-reasoning-schematic}). A typical approach to obtain the reasoning data for training is to annotate robot action trajectories with various foundation models. At training time, these reasoning texts are tokenized and prepended to the action tokens, which the model learns to generate via next-token prediction (same as for the base VLM's pre-training). Each inference step involves decoding the reasoning text followed by actions, then executing the latter, making ECoT policies' inference speeds low.

\vspace{-0.3cm}
\section{Why Does Embodied Chain-of-Thought Reasoning Improve Performance?}
\label{sec:hypotheses}

\vspace{-0.4cm}

We aim to develop more practical recipes for training embodied reasoning policies that avoid its conventional drawbacks. We start by formulating hypotheses for \emph{why} reasoning improves policy performance. We will then use these hypotheses to develop multiple novel, lightweight recipes for reasoning policy training in \cref{sec:ecot_lite_recipes}, and test them empirically in \cref{sec:experiments}.

\vspace{-0.4cm}

\paragraph{\textcolor{burntorange}{\textbf{Hypothesis 1: Embodied reasoning improves representation learning.}}}
One possible explanation for the benefit of reasoning is that the additional knowledge supplied to the model via the reasoning steps influences the model's representations. For example, the reasoning trace could indicate that a particular object's location is relevant. If improved representations were the primary explanation for the increase in policy performance, we would expect that actually generating reasonings at test-time is less important compared to supervising the model with the information contained in the reasoning steps, such that its internal representations are attentive to the features that the reasoning steps require it to predict. If this holds true, we can design approaches that use reasoning during training for improved representation learning, but forego slow reasoning at test time to retain low policy latency.

\vspace{-0.3cm}

\paragraph{\textcolor{bluegray}{\textbf{Hypothesis 2: Embodied reasoning provides learning curriculum.}}}
An alternative hypothesis is that embodied reasoning features provide the policy with an implicit learning curriculum, where it can first learn the comparatively straight-forward reasoning tasks, like mapping from low-level movement primitives to robot actions, and then ``work its way up'' to the full observation-to-action prediction task. Thus, the model may learn a more generalizable mapping from images to actions, instead of struggling from the start with the end-to-end prediction task and resorting to memorizing solutions. Similar effects have been observed in LLM training \citep{Kang24-learningdynamicsLLMgeneralization}. If this holds true, we can design training recipes where we simply use embodied reasoning as a ``scaffolding'' for training, but remove it entirely during inference to simplify and speed-up policy rollouts.

\vspace{-0.3cm}

\paragraph{\textcolor{asparagus}{\textbf{Hypothesis 3: Embodied reasoning increases effective model expressivity.}}}
Finally, embodied reasoning extends the \emph{length} of the sequence of tokens that the VLA is operating over. As such, the model leverages more compute during training and inference than comparably-sized, regular VLA policies, and thus effectively increases its expressiveness. To this end, multiple works in language and vision-language modeling have observed that merely increasing the \emph{number} of tokens, even without adding new information, can improve model performance \citep{Beyer24-paligemma, Steiner24-paligemma2, Pfau24-letsthinkdotdot, Goyal24-thinkBeforeYouSpeak}. Similarly, we may design a simplified training recipe that introduces additional tokens \emph{without} the need to perform extensive reasoning annotation on the data ahead of time.

\begin{figure}[t]
    \centering
    \includegraphics[width=\linewidth]{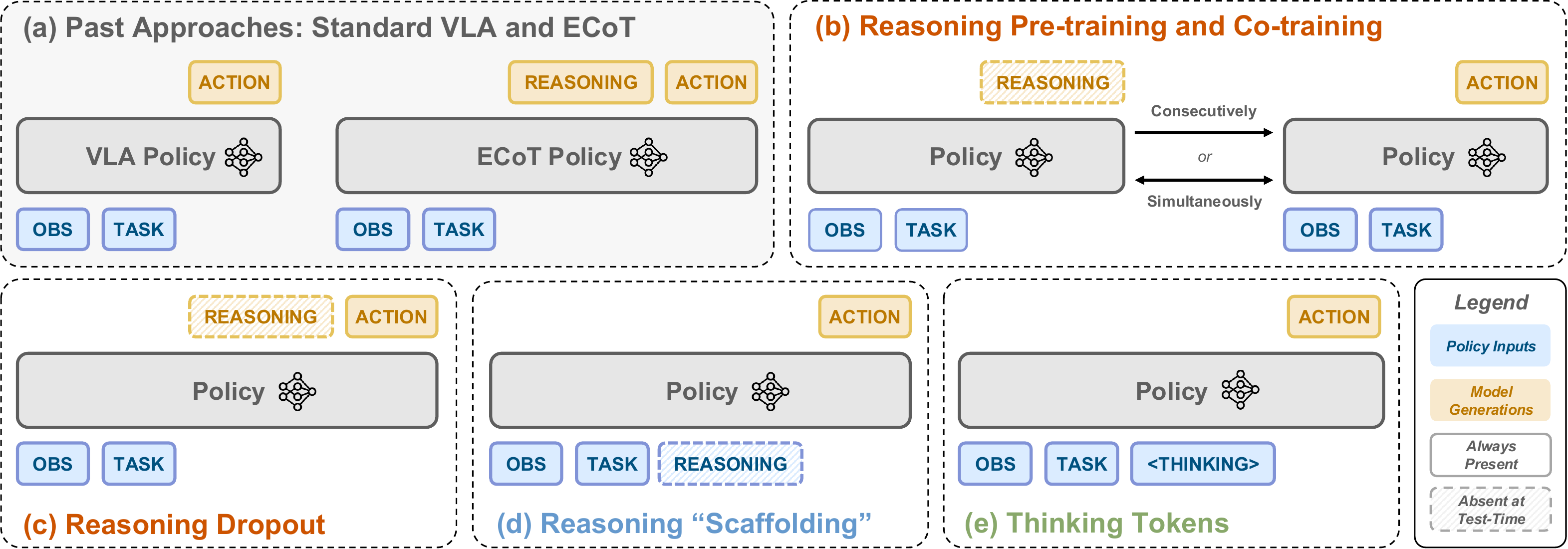}
    \vspace{-0.5cm}
    \caption{
        \Acronym{} training recipes. Blue indicates inputs, orange indicates outputs/generations, dashed border represents absence during test-time (and random drop-out during training). \textbf{(a)}: Standard VLA \citep{Brohan23-rt2, Kim24-openVLA} and embodied CoT \citep{Zawalski24-ecot} training. \textbf{(b)} Pre-train or co-train VLA models with embodied reasoning data. \textbf{(c)}: Provide reasoning data as a ``scaffolding'' in context during training. \textbf{(d)}: Train with reasoning dropout, remove reasoning during inference. \textbf{(e)}: Introduce non-semantic ``thinking tokens'' to increase effective model expressivity.
    }
    \vspace{-0.6cm}
    \label{fig:policy_variants}
\end{figure}

\section{\Acronym{}: Practical Training Recipes for Embodied Reasoning Policies}
\label{sec:ecot_lite_recipes}
\vspace{-0.4cm}

In this section, we use the hypotheses from the previous section to design a number of simplified embodied chain-of-thought training recipes. Our goal is to retain most of the generalization abilities of the original ECoT recipe introduced in \cref{sec:prelim}, but mitigate the need for either extensive data annotation or high-latency inference. We provide an overview of our proposed approaches in \cref{fig:policy_variants}.

\textbf{\textcolor{burntorange}{Reasoning pre- or co-training.}} Following \textcolor{burntorange}{\textbf{Hyp.~1}}, we can use the reasoning data to shape the policy's \emph{representations} by pre- or co-training the VLA on it -- both commonly used in other domains \citep{Gururangan20-dontstoppretraining, Lang22-cotrainingImprovesPromptBasedLearning, Brief24-mixingitup-cocktaileffect}. For pre-training, the base VLM is trained to just produce reasonings, then trained on just actions. For co-training, the VLM is trained on both objectives simultaneously -- half of the training batch maps observations to reasonings only, the other half to actions only. This treats reasoning as an auxiliary loss. At inference time, both pre-training and co-training policies operate as standard VLAs, without explicit reasoning and thus with low latency.

\textbf{\textcolor{burntorange}{Test-time reasoning dropout.}} Similar to the above approaches, the main goal here is to use embodied reasoning data to shape the policy's representation (\textcolor{burntorange}{\textbf{Hyp.~1}}). However, instead of treating reasoning as a fully separate auxiliary task, this approach trains the model to explicitly use the reasoning information \emph{for action prediction}, by \emph{sequencing} reasoning and action tokens. Importantly, it performs \emph{dropout} on the reasoning steps such that part of the training examples contain \emph{no} reasoning. This allows us to again use the policy without reasoning prediction at inference time, while explicitly using reasoning for action prediction during training may encourage better representational transfer than simple pre- or co-training. Thus, this approach allows us to isolate the impact of generating test-time reasonings by turning it on or off and measuring the resulting performance.

\textbf{\textcolor{bluegray}{Reasoning ``scaffolding''.}} As outlined in \textcolor{bluegray}{\textbf{Hyp.~2}}, we may want to use reasoning only as a scaffold for the learning process, \emph{without} requiring the policy to expend capacity for learning to actually predict the reasoning itself. We thus can provide reasoning examples \emph{in context} during training, but with no loss on them. We can again apply dropout on the reasoning steps such that a fraction of the training examples contain \emph{no} reasoning scaffold, i.e., train the policy to perform direct observation-to-action mapping. At inference time, we can then use this policy like a standard VLA, with low latency.

\textbf{\textcolor{asparagus}{Thinking tokens.}} We can increase the effective capacity of the policy, \emph{without} semantic reasoning, by introducing empty ``thinking tokens'' in the context of the model during both, training and inference (\textcolor{asparagus}{\textbf{Hyp.~3}}). In this way, we artificially increase the token sequence length and thus the computational resources at the VLA's disposal. Similar approaches have been explored in the context of LLM training~\citep{Pfau24-letsthinkdotdot, Goyal24-thinkBeforeYouSpeak}. We set the number of thinking tokens to a random value within the range of reasoning token lengths in our embodied reasoning data, to enable a fair, compute-matched comparison.

\vspace{-0.3cm}
\section{Experiments}
\label{sec:experiments}
\vspace{-0.3cm}

Our goal is to test the performance of our lightweight embodied reasoning recipes from \cref{sec:ecot_lite_recipes} in comparison to standard VLA and ECoT inference. Specifically, we aim to answer the following:

\vspace{-3pt}

\textbf{(1)} How does the performance of the \Acronym{} policies compare to regular VLAs and ECoT?
\vspace{-3pt}

\textbf{(2)} Which of the hypotheses in \cref{sec:hypotheses} is best supported by empirical results and explains why embodied reasoning improves policy generalization?
\vspace{-3pt}

\textbf{(3)} How can we decide \emph{which} embodied reasoning strategy to use in a given scenario?

\vspace{-0.3cm}
\subsection{Experimental Setup}
\label{sec:exp_setup}
\vspace{-0.2cm}

\begin{figure}[t]
    \centering
    \includegraphics[width=0.9\linewidth]{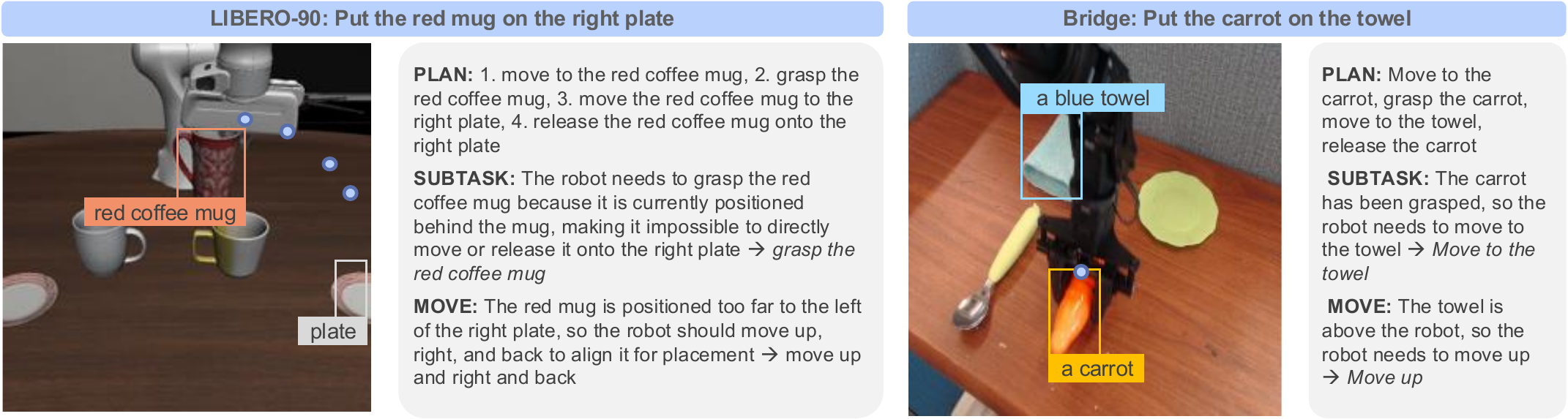}
    \caption{
        Example ECoT reasonings for LIBERO and Bridge. See \cref{fig:example-cots} for more examples.
    }
    \vspace{-0.5cm}
    \label{fig:example-cots-small}
\end{figure}

\textbf{Environments \& Tasks.} We evaluate our approach in two environments: the widely-used, simulated LIBERO robot manipulation benchmark~\citep{Liu23-libero}, and a similar distribution of evaluation tasks from the BridgeData~V2 ~\citep{Walke23-bridgeDataV2} environment that was used in prior works on VLAs and robot reasoning \citep{Kim24-openVLA, Zawalski24-ecot}. In both environments, we test on a diverse set of tasks that requires policies to generalize beyond their training data: in LIBERO we introduce multiple levels of randomization to initial object positions and distractor objects; in the BridgeData~V2 evaluations we follow~\citet{Zawalski24-ecot} and~\citet{Kim24-openVLA} and evaluate on in-distribution, motion generalization, spatial relations, and unseen objects. For both environments we train policies on the publicly available LIBERO and Bridge~V2 training datasets. For more details on robot setups, datasets, and tasks, see~\cref{sec:app:environment_details} or \cref{fig:example-cots-small} for reasoning examples. We refer to \cref{sec:app:reasoning_data} for details of how the embodied reasoning annotations are created.

\textbf{Policy training.}
For each environment, we use standard, auto-regressive VLA architectures (\cref{sec:prelim}) from prior work (see \cref{sec:app:policy_details} for details).
We emphasize that our policy architecture choices are merely trying to mirror prior state-of-the-art policies on the respective environments, but the embodied reasoning training recipes we develop are \emph{agnostic} to the choice of the underlying VLA and can readily be applied to other VLA architectures.

\textbf{Comparisons.} We compare our \Acronym{} training recipes to standard (non-reasoning) VLA training \citep{Kim24-openVLA, Belkhale24-miniVLA}, and prior embodied chain-of-thought policy training approaches \citep{Zawalski24-ecot}. We ensure that all approaches are trained on the same robot demonstration and reasoning data, using the same amount of computational resources, and evaluated under comparable initial states, lighting conditions, and camera angles. In total, our results contain 121,500 simulated and 444 real-robot trials.

\vspace{-0.2cm}
\subsection{\Acronym{} Enables Generalizable Policies with Fast Inference}
\label{sec:results}
\vspace{-0.3cm}

\begin{figure}[t]
    \centering
    \includegraphics[width=\linewidth]{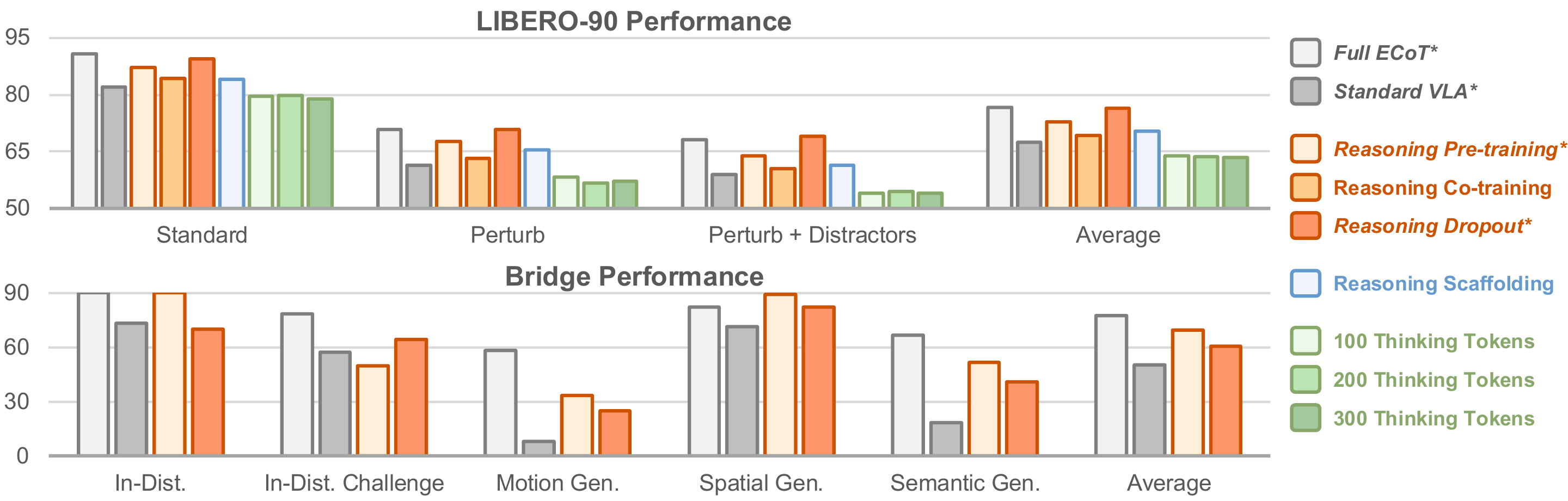}
    \vspace{-0.5cm}
    \caption{
        \footnotesize{\textbf{Top:} Performance of all methods on LIBERO-90 benchmarks. The most performant approaches are ECoT and the \Acronym{} reasoning dropout policy, both of which beat past state-of-the-art on the standard LIBERO-90 evaluations (90.8\% and 89.4\% vs. 88.6\% by \citet{Mete24-quest}). Reasoning pre-training also improves performance significantly. See \cref{tab:libero-results} for numerical values and standard errors.
        \textbf{Bottom:} We replicate the reasoning dropout and pre-training policies in Bridge to validate their real-world effectiveness. Both \Acronym{} approaches improve on the standard VLA's performance. While full ECoT is the most performant, the \Acronym{} policies do not generate test-time reasonings, making their inference speeds much faster. See \cref{tab:bridge-results} for per-task numerical values and standard errors. Asterisks in legend indicate method appears in top and bottom.}
    }
    \vspace{-0.6cm}
    \label{fig:results}
\end{figure}

We present our LIBERO policies' performance in \cref{fig:results} (top). ECoT and reasoning dropout both significantly increases performance in all three LIBERO-90 splits over the equivalent standard MiniVLA policy. Both achieve around 90\% on standard LIBERO-90, \textbf{exceeding past LIBERO-90 state-of-the-art results by \citet{Mete24-quest} (88.6\%)}. Critically, reasoning dropout does not generate test-time reasonings, so it is much faster than full ECoT. The reasoning pre-training policy yields the second biggest increase in performance over the standard VLA (+5.4\%), while co-training and scaffolding yield smaller improvements. No thinking token condition improves over the baseline.

As reasoning dropout and pre-training were the most effective approaches in LIBERO, we validate their real-world applicability in Bridge tasks, presented in \cref{fig:results} (bottom). Both approaches improve significantly over the standard VLA baseline, and while ECoT remains the most performant, our novel approaches are around 3$\times$ faster. Unlike with LIBERO, we find that reasoning dropout is \textit{less} effective than reasoning pre-training for Bridge. We now discuss these results in more detail.

\subsection{Representation Learning Results}
\label{subsec:representation-learning-results}
\vspace{-0.3cm}

All our representation learning policies improve on the non-reasoning VLA, in support of \textcolor{burntorange}{\textbf{Hyp.~1}}. We find that reasoning co-training only makes a marginal improvement over the baseline (+1.9\%), but pre-training makes a more significant one (+5.4\%). \citet{Gao25-starGen-taxonomy} similarly notes a mixed impact of co-training on actions and VQA data. Our result is perhaps more surprising, as we are co-training on data from the same domain as our action prediction task, while VQA data is generally more distinct from robot data. Indeed, the fact that reasoning pre-training, dropout, or standard ECoT all significantly improve performance shows that learning from the reasoning data \textit{should} be helpful for improving action prediction; co-training just seems to be an ineffective way to do so.

\textbf{Why is pre-training better than the baseline?} Reasoning pre-training causes the model's internal representations to capture the reasoning features that are helpful for robust action prediction. In contrast, tuning a base VLM to predict actions in the standard VLA recipe may cause the resulting policy to learn a poor mapping, as it has to leverage representations learned from \textit{general} vision-language data (which may be less suited to action prediction, due to low exposure to robot data). This in turn leads to lower performance. Such an argument is commonly applied to adapting LLMs \citep{Gururangan20-dontstoppretraining}.

\textbf{Why is co-training worse than pre-training?} When co-training, the policy does still learn to generate reasonings, so its representations should also be improved. However, it critically learns these \textit{at the same time} that it fits a mapping from observations to actions. We suspect that, once learning to reason induces good representations, the VLA has already learned a poor mapping from images to actions, i.e., the model learns to predict actions and reasonings with separate parameters (with little overlap). At this point, it may be hard for action prediction to ``switch'' to being based on the good reasoning representations. In contrast, by having reasoning as the sole pre-training objective, the model devotes all parameters to modeling the features needed to reason; when action fine-tuning starts, the model \textit{must} tune parameters that were used for reasoning. Similar arguments have been presented in past works \citep{Ren23-prepareTaskHeadForFinetuning}, which we discuss in more detail in \cref{app_subsec:pretrain-vs-cotrain}.

\textbf{When does generating test-time reasonings matter?} We find that our reasoning dropout policy performs best on LIBERO-90 \textit{regardless} of if it generates test-time reasonings. We suspect this may be a byproduct of LIBERO-90 being narrow (containing 90 tasks vs. the thousands in real robot datasets like Bridge or DROID \citep{Ebert21-bridgeDataV1, Walke23-bridgeDataV2, Khazatsky24-droid}). As a result, the LIBERO reasoning steps are not very diverse; features like the plan, subtasks, and object labels have little variation for any given task. Thus, it is likely much easier for the model to fully memorize and ``internalize'' the reasonings, making generating them at test time superfluous. This is corroborated by our more diverse Bridge evaluation, wherein enabling test-time reasoning \textit{does} improve performance compared to disabling it. Qualitatively, this performance gap mainly comes from the reasoning dropout policy picking up incorrect objects or colliding with obstacles -- failure modes that would intuitively be alleviated by generating bounding boxes or motion rationales in test-time reasonings (see \cref{fig:failure-modes} for examples).

\vspace{-0.2cm}
\subsection{Learning Curriculum Results}
\vspace{-0.2cm}

Our reasoning scaffolding approach yields slightly better performance than the baseline (+2.9\%), comparable to reasoning co-training. This weakly supports \textcolor{bluegray}{\textbf{Hyp.~2}}, that having reasonings in-context during training \textit{does} improve the learned observation to action mapping, even when reasonings are absent at test time. Notably, this approach only differs from reasoning dropout in that the latter also assigns losses to reasoning tokens; it is therefore the combination of scaffolding \textit{and} learning to reason that leads to the most significant improvement for LIBERO.

\vspace{-0.2cm}
\subsection{Improved Expressivity Results}
\vspace{-0.2cm}

We find that adding thinking tokens degrades performance relative to the baseline (-3.8\% on average). This runs counter to \textcolor{asparagus}{\textbf{Hyp.~3}} and empirical results in language modeling, wherein thinking tokens improve performance in various domains \citep{Pfau24-letsthinkdotdot, Goyal24-thinkBeforeYouSpeak}, suggesting that the main benefit of robot reasoning is \textit{not} increased policy expressivity, but in actually learning to produce semantically-meaningful reasoning steps.
We note that these experiments use MiniVLA \citep{Belkhale24-miniVLA}, which is much smaller than other popular VLAs (1B vs. 55B for RT-2, 7B for OpenVLA, and 3B for $\pi_0$ \citep{Brohan23-rt2, Kim24-openVLA, Black24-pi0}).
Our results nonetheless indicate that expressivity is not a bottleneck for it.

\vspace{-0.3cm}
\section{Which Robot Reasoning Approach is Best for My Problem?}
\vspace{-0.3cm}

As shown in \cref{fig:results}, we find that when trained on the same demonstration data, standard VLAs are universally outperformed by ECoT and our two \Acronym{} variants, indicating that all three successfully use the same embodied reasoning datasets to improve policy performance. To determine which one is appropriate, we thus now consider their compute, inference, and performance tradeoffs, then make prescriptions for when to use each.

First, we find that \textbf{embodied chain-of-thought reasoning} is the most performant approach. However, it is also the slowest: with TensorRT-LLM FP8 compilation on an H100 GPU \citep{Nvidia-tensorRT-LLM, Chen25-tensorrt-openvla}, the policy only achieves around 1-1.2 Hz control frequency (or 0.3-0.5 Hz on 4090 without compilation), compared to VLAs of the same architecture, which achieve around 3-4 Hz (4090, no compilation\footnote{Most of the compilation speed-up comes from FP8, which must be run on Hopper GPUs (it is unsupported on 4090) \citep{Chen25-tensorrt-openvla}. We also find that TensorRT-LLM compilation does not speed up non-reasoning policies much.}). As both \Acronym{} variants also map observations directly to actions, they share that faster inference speed.

Of these variants, \textbf{reasoning dropout} seems better in ``narrow'' task domains, matching the state-of-the-art performance of full ECoT on all variants of LIBERO (\cref{fig:results}, top). As it is trained the same way as ECoT (just with the reasoning occasionally dropped out), dropout has the same resource demands as ECoT, and even affords users the flexibility of ``turning on'' the reasonings at test time.

Finally, \textbf{reasoning pre-training} does better than dropout on Bridge in all but one split (\cref{fig:results}, bottom). It has the downside of needing more gradient steps in order to learn reasoning and action prediction consecutively (whereas ECoT and reasoning dropout learns them simultaneously). However, this also means that it does not need \textit{paired} reasoning steps and action data\footnote{In principle, it can learn from arbitrary embodied robot reasoning data, including from other embodiments. However, we leave investigations into this possibility to future works.}. Last, as it forgoes training on reasonings and actions in the same context window, each training datapoint requires less memory, which is particularly salient when actions are expressed as many tokens.

\textbf{Our prescription is:} use full ECoT to maximize performance, at the cost of slower inference. Use reasoning dropout in narrower task domains or if the option to ``turn on'' test-time robot reasonings is needed. Use reasoning pre-training in more diverse task domains, or if unpaired embodied reasoning data is available (and training for more steps is acceptable).

\vspace{-0.3cm}
\section{Discussion}
\label{sec:discussion}
\vspace{-0.3cm}

We investigated three hypotheses for why chain-of-thought reasoning improves learned robot policies, namely robot reasonings aid representation learning, provide learning curricula, or increased policy expressivity. By isolating each of these improvement mechanisms in extensive simulated experiments, we find that reasoning pre-training and test-time reasoning dropout are effective ways for maintaining the improved performance conferred by robot reasoning while avoiding its slower inference speeds. We validate these approaches' effectiveness in real-world manipulation experiments, then finally give explicit prescriptions for when each approach is appropriate.

\clearpage
\section{Limitations}

While we have shown the effectiveness of our proposed approaches, they do not address all of robot reasonings' limitations. Our results show that policy expressivity is likely \textit{not} a bottleneck for VLAs, with our thinking token approach failing to improve performance. While unsurprising, this means that to enjoy the benefits of robot reasoning, users need robot reasoning training data, which can be difficult or expensive to extract.

Additionally, while we have conducted extensive empirical analysis on our proposed approaches' \textit{performance} and have provided intuitive speculation on why they are or are not effective, we leave investigation of reasoning's impact on actual model learning dynamics to future works. Such fine-grained analysis may prove useful for understanding \textit{why} certain training schemes yield better representational transfer or grounding than others.

Lastly, as part of our analysis, we hold many design choices (e.g., policy architecture, training hyperparameters, and reasoning corpora) constant to make our comparisons more controlled and fair. While we have presented empirical evidence of \Acronym's performance benefits, further optimizations are possible -- for example, for reasoning pre-training, one could investigate if said approach enables better cross embodiment \textit{reasoning} transfer, taking advantage of how said approach does not need paired reasoning-action data.

\acknowledgments{We thank Aviral Kumar, Evan Hernandez, Kyle Stachowicz, Seohong Park, and Vivek Myers for insightful discussions. This research was partly supported by ONR N00014-25-1-2060, NSF IIS-2150826, and ARL DCIST CRA W911NF-17-2-0181, with additional support from Volkswagen and NVIDIA. We also thank the NVIDIA Academic Grant Program for providing the compute resources used in this work.}

\bibliography{example}  % .bib

\clearpage
\appendix
\input{appendix.tex}

\end{document}

%% file: appendix.tex
\section{Full Results and Further Discussion}

\begin{table}[]
\caption{Performance of all proposed approaches and baselines on all Libero-90 benchmark variants (Mean $\pm$ StdErr). Success rates are computed for 50 trials on all 90 tasks (4500 episodes per variant, or 13500 total episodes per policy).}
\label{tab:libero-results}
\resizebox{\textwidth}{!}{%
\begin{tabular}{@{}cccccccccc@{}}
\toprule
\multirow{2}{*}{\textbf{\begin{tabular}[c]{@{}c@{}}Libero-90\\ Variant\end{tabular}}} &
  \multicolumn{2}{c|}{\textbf{Baselines}} &
  \multicolumn{3}{c|}{\textbf{\textcolor{burntorange}{Representation Learning}}} &
  \multicolumn{1}{c|}{\textbf{\textcolor{bluegray}{Learning Curriculum}}} &
  \multicolumn{3}{c}{\textbf{\textcolor{asparagus}{Improved Expressivity}}} \\
 &
  \textbf{Full ECoT} &
  \multicolumn{1}{c|}{\textbf{Standard VLA}} &
  \textbf{\begin{tabular}[c]{@{}c@{}}Reasoning\\ Pre-training\end{tabular}} &
  \textbf{\begin{tabular}[c]{@{}c@{}}Reasoning\\ Co-training\end{tabular}} &
  \multicolumn{1}{c|}{\textbf{\begin{tabular}[c]{@{}c@{}}Reasoning\\ Dropout\end{tabular}}} &
  \multicolumn{1}{c|}{\textbf{\begin{tabular}[c]{@{}c@{}}Reasoning\\ Scaffolding\end{tabular}}} &
  \textbf{\begin{tabular}[c]{@{}c@{}}100 Thinking\\ Tokens\end{tabular}} &
  \textbf{\begin{tabular}[c]{@{}c@{}}200 Thinking\\ Tokens\end{tabular}} &
  \textbf{\begin{tabular}[c]{@{}c@{}}300 Thinking\\ Tokens\end{tabular}} \\ \midrule
\textbf{Standard} &
  \textbf{90.8\%} &
  82.0\% &
  \textit{87.1\%} &
  84.2\% &
  \textbf{89.4\%} &
  84.1\% &
  79.5\% &
  79.8\% &
  78.9\% \\ \cmidrule(r){1-1}
\textbf{Perturb} &
  \textbf{70.8\%} &
  61.4\% &
  \textit{67.6\%} &
  63.1\% &
  \textbf{70.8\%} &
  65.4\% &
  58.3\% &
  56.8\% &
  57.2\% \\ \cmidrule(r){1-1}
\textbf{\begin{tabular}[c]{@{}c@{}}Perturb +\\ Distractors\end{tabular}} &
  \textbf{68.2\%} &
  58.9\% &
  \textit{63.9\%} &
  60.5\% &
  \textbf{69.1\%} &
  61.3\% &
  54.0\% &
  54.4\% &
  54.0\% \\ \midrule
\textbf{Average} &
  \textbf{\begin{tabular}[c]{@{}c@{}}76.6\%\\ $\pm$ 0.36\%\end{tabular}} &
  \begin{tabular}[c]{@{}c@{}}67.4\%\\ $\pm$ 0.40\%\end{tabular} &
  \textit{\begin{tabular}[c]{@{}c@{}}72.8\% \\ $\pm$ 0.38\%\end{tabular}} &
  \begin{tabular}[c]{@{}c@{}}69.3\% \\ $\pm$ 0.40\%\end{tabular} &
  \textbf{\begin{tabular}[c]{@{}c@{}}76.4\% \\ $\pm$ 0.37\%\end{tabular}} &
  \begin{tabular}[c]{@{}c@{}}70.3\% \\ $\pm$ 0.39\%\end{tabular} &
  \begin{tabular}[c]{@{}c@{}}63.9\% \\ $\pm$ 0.41\%\end{tabular} &
  \begin{tabular}[c]{@{}c@{}}63.7\% \\ $\pm$ 0.41\%\end{tabular} &
  \begin{tabular}[c]{@{}c@{}}63.4\% \\ $\pm$ 0.41\%\end{tabular} \\ \bottomrule
\end{tabular}%
}
\end{table}

\begin{table}[h]
\centering
\caption{Per-task performance of the best \Acronym{} variants, replicated in Bridge and compared against ECoT and standard VLA baselines.}
\label{tab:bridge-results}
\resizebox{0.9\textwidth}{!}{%
\begin{tabular}{@{}clcccc@{}}
\toprule
\multirow{2}{*}{\textbf{Split}} &
  \multicolumn{1}{c}{\multirow{2}{*}{\textbf{Task}}} &
  \multicolumn{2}{c|}{\textbf{Past Methods}} &
  \multicolumn{2}{c}{\textbf{\Acronym{} Variants}} \\
 &
  \multicolumn{1}{c}{} &
  \textbf{\begin{tabular}[c]{@{}c@{}}Full\\ ECoT\end{tabular}} &
  \multicolumn{1}{c|}{\textbf{\begin{tabular}[c]{@{}c@{}}Standard\\ VLA\end{tabular}}} &
  \textbf{\begin{tabular}[c]{@{}c@{}}Reasoning\\ Pre-training\end{tabular}} &
  \textbf{\begin{tabular}[c]{@{}c@{}}Reasoning\\ Dropout\end{tabular}} \\ \midrule
\multirow{2}{*}{\textbf{In Dist.}} &
  \begin{tabular}[c]{@{}l@{}}Put the {[}mushroom / corn / \\ eggplant{]} in the {[}pot / bowl{]}\end{tabular} &
  \textbf{88.9\%} &
  72.2\% &
  \textbf{88.9\%} &
  72.2\% \\
 &
  \begin{tabular}[c]{@{}l@{}}Place the {[}spoon / carrot{]} in\\ on the {[}towel / plate{]}\end{tabular} &
  \textbf{91.7\%} &
  75.0\% &
  \textbf{91.7\%} &
  66.7\% \\ \midrule
\multirow{2}{*}{\textbf{\begin{tabular}[c]{@{}c@{}}In Dist.\\ Challenge\end{tabular}}} &
  \begin{tabular}[c]{@{}l@{}}Put the {[}broccoli / spoon / \\ cube{]} on the towel\\ (all green objects)\end{tabular} &
  \textbf{83.3\%} &
  16.7\% &
  66.7\% &
  66.7\% \\
 &
  \begin{tabular}[c]{@{}l@{}}Put the {[}green / pink{]} spoon\\ on the {[}plate / towel{]}\end{tabular} &
  75.0\% &
  \textbf{87.5\%} &
  37.5\% &
  62.5\% \\ \midrule
\textbf{\begin{tabular}[c]{@{}c@{}}Motion\\ Gen.\end{tabular}} &
  \begin{tabular}[c]{@{}l@{}}Put the {[}carrot / mushroom{]}\\ on the {[}plate / pot{]}\\ (target location is high up)\end{tabular} &
  \textbf{58.3\%} &
  8.3\% &
  33.3\% &
  25.0\% \\ \midrule
\multirow{3}{*}{\textbf{\begin{tabular}[c]{@{}c@{}}Spatial\\ Gen.\end{tabular}}} &
  \begin{tabular}[c]{@{}l@{}}Put the {[}banana / tomato{]} in\\ the {[}left / right{]} bowl\end{tabular} &
  \textbf{91.7\%} &
  \textbf{91.7\%} &
  83.3\% &
  75.0\% \\
 &
  \begin{tabular}[c]{@{}l@{}}Put the {[}green / orange{]} toy\\ in the {[}left / right{]} pot\end{tabular} &
  75.0\% &
  50.0\% &
  \textbf{87.5\%} &
  75.0\% \\
 &
  \begin{tabular}[c]{@{}l@{}}Move the {[}mushroom / carrot{]}\\ to the {[}left / right{]} of the\\ {[}carrot / mushroom{]}\end{tabular} &
  75.0\% &
  62.5\% &
  \textbf{100\%} &
  \textbf{100\%} \\ \midrule
\multirow{4}{*}{\textbf{\begin{tabular}[c]{@{}c@{}}Semantic\\ Gen.\end{tabular}}} &
  \begin{tabular}[c]{@{}l@{}}Put the watermelon on \\ the towel\end{tabular} &
  66.7\% &
  0\% &
  \textbf{83.3\%} &
  83.3\% \\
 &
  Put the toothbrush on the plate &
  66.7\% &
  33.3\% &
  \textbf{83.3\%} &
  50.0\% \\
 &
  Put the screw in the bowel &
  \textbf{66.7\%} &
  33.3\% &
  \textbf{66.7\%} &
  16.7\% \\
 &
  \begin{tabular}[c]{@{}l@{}}Reach for the {[}ketchup / wrench\\ / mallet{]}\end{tabular} &
  \textbf{66.7\%} &
  11.1\% &
  0\% &
  22.2\% \\ \midrule
\multicolumn{2}{c}{\textbf{Aggregate}} &
  \textbf{77.5\% $\pm$ 4.0\%} &
  50.5\% $\pm$ 4.7\% &
  69.4\% $\pm$ 4.4\% &
  60.4\% $\pm$ 4.6\% \\ \bottomrule
\end{tabular}%
}
\end{table}

\begin{figure}[t]
    \centering
    \includegraphics[width=\linewidth]{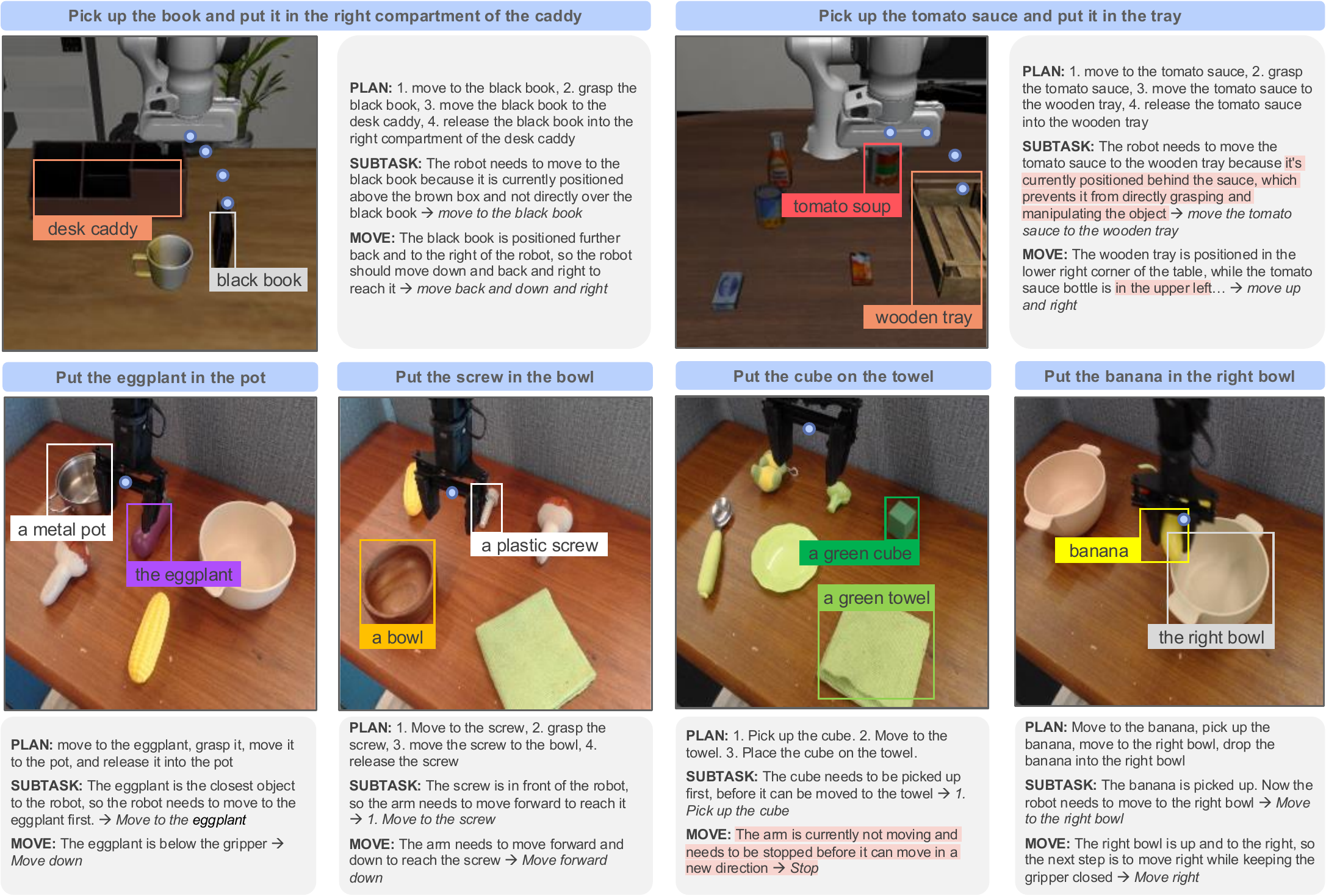}
    \vspace{-0.5cm}
    \caption{
        \footnotesize{More examples of ECoT reasoning steps generated by our policies (LIBERO on top, Bridge on bottom). Red highlights indicate incorrect or hallucinated features (though they still lead to task successes). Note the slight stylistic differences in the LIBERO reasonings, as the training data generation pipeline for that domain uses a different set of foundation models.}
    }
    \vspace{-0.3cm}
    \label{fig:example-cots}
\end{figure}

We provide numerical performance values for all our simulated LIBERO and real-world Bridge evaluations in \cref{tab:libero-results} and \cref{tab:bridge-results} respectively. These are identical to the values shown in \cref{fig:results}, just written as explicit success rate values (and divided per-task for the Bridge evaluations).

\subsection{Further Discussion: Pre-training vs. Co-training}
\label{app_subsec:pretrain-vs-cotrain}

\begin{figure}[t]
    \centering
    \includegraphics[width=\linewidth]{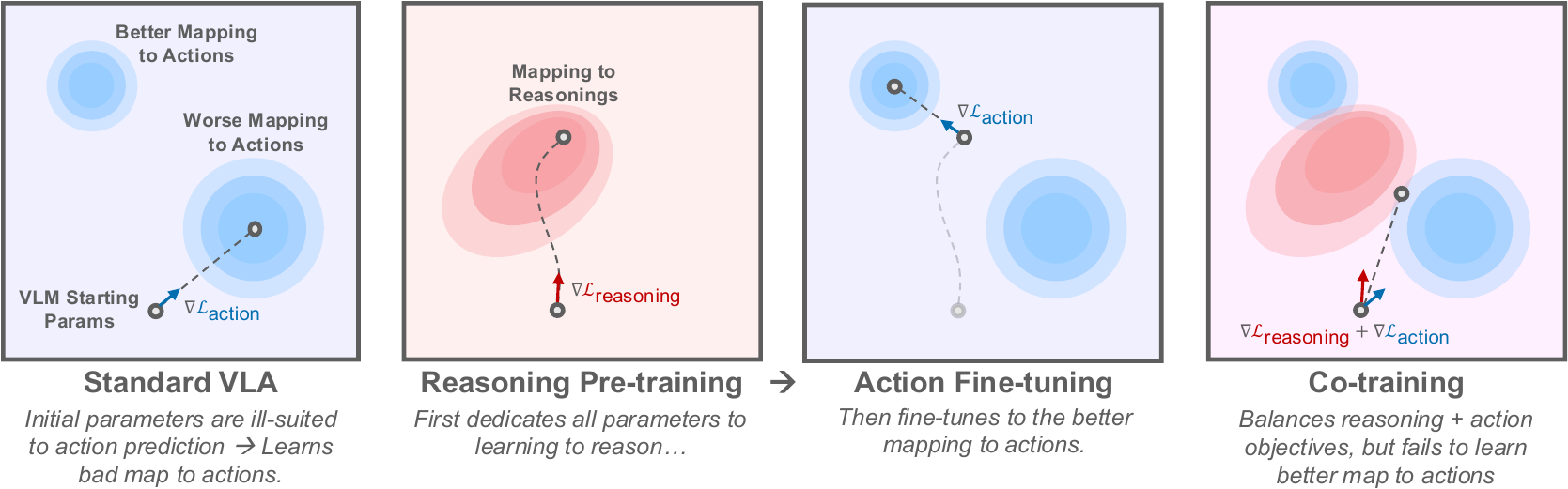}
    \vspace{-0.5cm}
    \caption{
        \footnotesize{Very abstract illustration of our argument as to why reasoning pre-training seems more effective than co-training, despite using the same data. \textcolor{blue}{Blue} indicates the loss landscape of the action prediction task, \textcolor{red}{red} corresponds to that of reasoning, and darker is lower loss in both cases. Co-training linearly mixes these two loss landscapes and aims to optimize both simultaneously, while pre-training optimizes reasoning and actions disjointly and consecutively. The latter seems to find a better mapping from observations to actions than the former, as shown by our LIBERO results. We suspect this is from the model dedicating all of its parameters and representational capacity to learning reasoning when pre-training, which leads to a part of parameter space that makes learning good actions easier. Note that we do not illustrate the loss landscapes of any approach wherein actions can attend to reasonings (dropout, scaffolding, or ECoT). In that case, since the representations of actions depend on the representations of reasonings, the overall loss landscape is not merely a linear combination of the two tasks' separate landscapes, meaning it is not easily related to the above abstraction.}
    }
    \vspace{-0.5cm}
    \label{fig:loss-landscape}
\end{figure}

We provide a (heavily abstracted) visualization of our argument as to why reasoning co-training seems to work poorly compared to pre-training in \cref{fig:loss-landscape}. 

Looking at past LLM literature, both pre-training \citep{Gururangan20-dontstoppretraining} and co-training \citep{Brief24-mixingitup-cocktaileffect, Lang22-cotrainingImprovesPromptBasedLearning} on task domain data have been shown to empirically boost model performance. We note that, in these co-training works, the co-training dataset is typically quite large, whereas our robot embodied reasoning datasets are not. If our hypothesis that co-training causes the model to learn action prediction and reasoning separately for LIBERO, it is possible that scaling up reasoning data (e.g., introducing more embodied tasks \textit{or} more robot embodiments) would force more parameter sharing between the two tasks, leading to better transfer.

While not in the context of co- vs. pre-training, several past works have investigated when fine-tuning is effective. For instance, while \citet{Ren23-prepareTaskHeadForFinetuning} investigate the setting of how to best initialize a model head atop a pre-trained backbone, they give a somewhat similar argument as what we give in \cref{subsec:representation-learning-results}. Specifically, they posit if a model head is over-tuned (i.e., to near minimal loss) while the pre-trained backbone is frozen, it could degrade downstream fine-tuning and performance -- paralleling our argument that learning action prediction in tandem with reasoning leads to the former getting stuck in a poor-performance solution.

Our argument also connects to the work of \citet{Kumar22-finetuningDistortsPretrainedFeatures} in a nuanced way. The authors show how, if a model's pre-trained representations are ``good'' for a downstream task, excessive fine-tuning can destroy those representations, lowering its performance on out-of-distribution inputs. They find that an effective alternative is to only fine-tune a task-specific linear head on the task of interest, thereby both (1) enforcing a simplicity bias when adapting and (2) better preserving the base model's representations (both of which lead to better performance). Connecting this to our results, co-training serves to ``preserve'' the reasoning representations throughout training, while reasoning pre-training followed by action fine-tuning does not. Given that the reasoning pre-training policy matches or outperforms even the full ECoT policy in all but one of the semantic generalization tasks, it seems that any distortion of pre-trained features induced by action fine-tuning is not very impactful. Rather, what is important is that the action predictions are grounded in the reasoning representations -- preserving them does not matter if the actions do not depend strongly on them in the first place.

\subsection{Further Discussion: Dropout vs. Full ECoT}
\label{app_subsec:dropout-vs-ecot}

\begin{figure}[t]
    \centering
    \includegraphics[width=\linewidth]{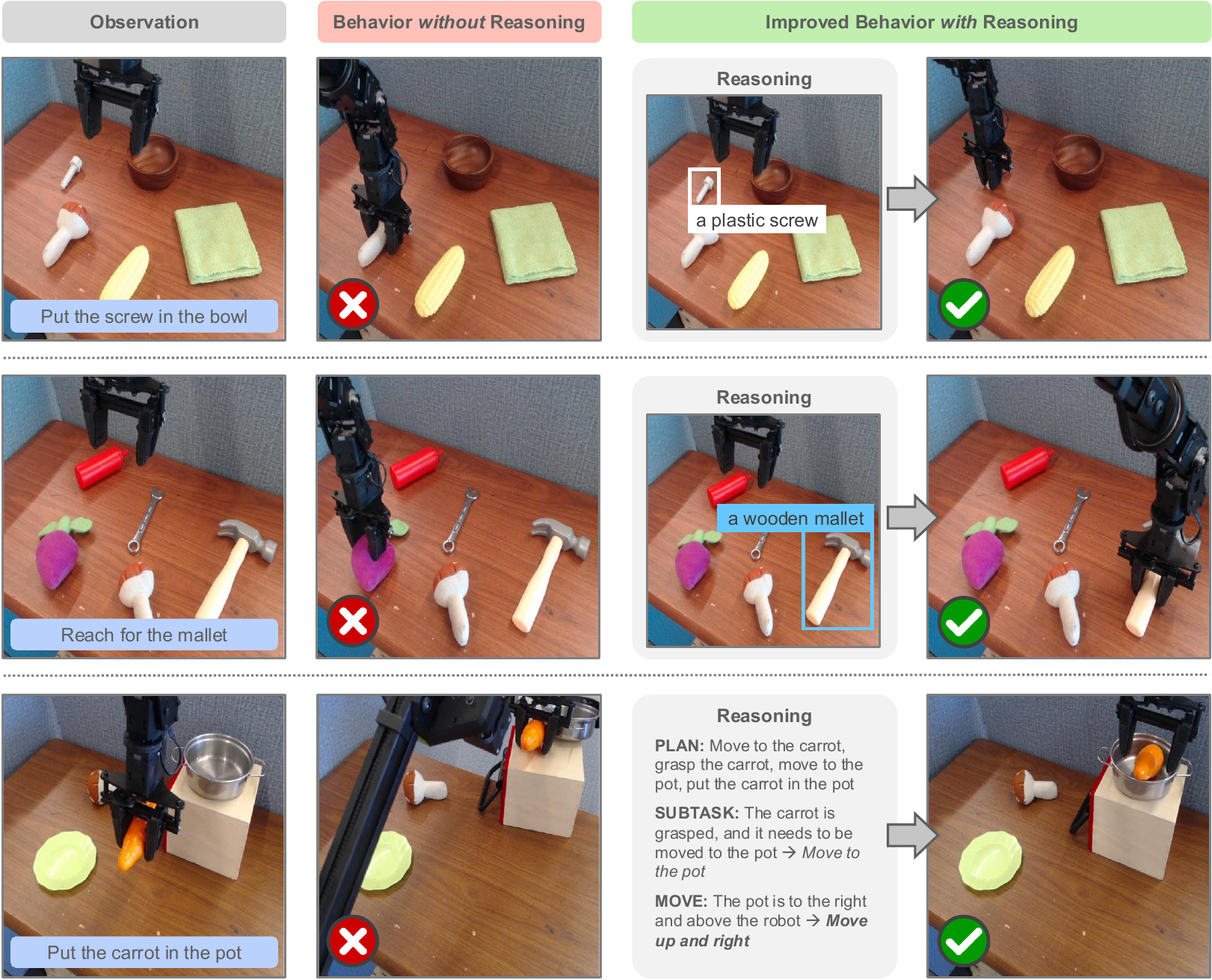}
    \vspace{-0.5cm}
    \caption{
        \footnotesize{Qualitative examples of the importance of test-time robot reasonings. We show policy behaviors on three different Bridge tasks with the reasonings disabled (reasoning dropout) or enabled (full ECoT), as well as the parts of the reasoning that intuitively lead to correct behaviors. The former leads to failure, while the latter leads to success. In the top and middle tasks, the target grasp object is out-of-distribution. However, the reasoning policy succeeds in labeling it with a bounding box, leading to correct grasping. At the bottom, disabling reasoning causes the robot to collide with the platform and pot, while enabling it causes the arm to move sufficiently high.}
    }
    \vspace{-0.3cm}
    \label{fig:failure-modes}
\end{figure}

We provide examples of when enabling reasoning results in better performance in \cref{fig:failure-modes}. These examples are drawn from the motion generalization and semantic generalization splits of our Bridge evaluations (\cref{tab:bridge-results}). The reasonings contain steps that intuitively help the robot choose good actions. When the reasonings are disabled (in reasoning dropout) or completely absent (in the standard VLA recipe), the policy tends to fail at these tasks.

Disabling reasonings does not see to affect the spatial generalization tasks (which test the policies' ability to understand spatial relations, listed as the fourth split in \cref{tab:bridge-results}). We suspect this comes from how Bridge containing numerous examples of left/right directional references in its task labels, allowing non-reasoning policies to ``internalize'' those concepts.

Curiously, we find that reasoning dropout does very well in the semantic generalization tasks (except the reaching ones).

\section{Environment Details}
\label{sec:app:environment_details}

% KP: this section should explain both Libero and Bridge setups. In detail, explain robot setups (observation + action spaces), training datasets (not about reasoning annotations), and evaluation tasks incl full list of prompts.
% KP: extra points for showing some initial states and the variation in those states

\subsection{LIBERO-90 Description}
\label{app_subsec:libero-description}

\begin{figure}[]
    % \centering
    % \includegraphics[width=\linewidth]{figures/chatgpt_intervention_prompt.jpeg}
    {\tiny\texttt{\input{text_figs/LiberoTasks1.txt}}}
    \caption{
        \footnotesize{First half of the 90 task prompts for LIBERO-90. Note some are repeated in different scenes (and thus they appear multiple times in this list)}
    }
    \label{fig:libero-tasks-1}
\end{figure}

\begin{figure}[]
    % \centering
    % \includegraphics[width=\linewidth]{figures/chatgpt_intervention_prompt.jpeg}
    {\tiny\texttt{\input{text_figs/LiberoTasks2.txt}}}
    \caption{
        \footnotesize{Latter half of the 90 task prompts for LIBERO-90. Note some are repeated in different scenes (and thus they appear multiple times in this list)}
    }
    \label{fig:libero-tasks-2}
\end{figure}

LIBERO-90 provides a suite of 90 simulated manipulation benchmarking tasks. The platform has both standardized evaluation features (episode spawn ranges, objects, scenes, success detection, controllable random seeds, etc) and a demonstration dataset of 50 rollouts per episode \citep{Liu23-libero}. We use the latter to train our LIBERO policies via behavioral cloning, as many past works have done \citep{Black24-pi0, Kim25-openVLA-oft, Mete24-quest, Belkhale24-miniVLA}. We note that we filter out any trajectories that, when rolled out, do not lead to successes, leaving us with 3,917 of the total 4,500 trajectories. Additionally, we emphasize that this is the only demonstration data we train upon. As it is drawn from the default episode distribution, that makes our challenge splits (described in more detail below) a strict distribution shift from the training data, thereby testing our policies' generalization.

In evaluating different policies, we use the same episode random seeds for each one, thereby ensuring that they are tested on the same set of initial conditions. Our evaluation code builds upon that of \citet{Belkhale24-miniVLA}, which starts episodes with 10 no-op steps (to allow spawned objects to settle) and ends them in failure after 400 more steps have passed without succeeding. The observation and action spaces for this environment are the standard ones: the former just contains RGB images from a fixed, third-person camera, while the latter is a 7D vector (6 joint angles and one for gripper open/close). The environment also provides proprioception, depth images, and wrist camera views as part of the observation space. While these have been shown to improve performance (including for VLAs \citep{Belkhale24-miniVLA}), we do not use them, as our goal is to determine the impacts of robot reasoning on VLA performance, \textit{not} to optimize our systems for LIBERO in particular. We emphasize that this is a design choice, and not a core limitation of any of our approaches. We provide the full list of 90 tasks in \cref{fig:libero-tasks-1} and \cref{fig:libero-tasks-2}.

\subsection{Custom Challenge Split Details}
\label{app_subsec:libero-challenge}

We develop a custom challenge split for the existing Libero-90 environments in order to test policy generalization. We introduce two types of randomization: \textit{Perturbation} and \textit{Distractors}. Perturbations involve expanding the initial state distribution in which task-relevant objects are randomized at the beginning of each episode. Task specifications for the standard Libero-90 tasks specify a rectangular initial distribution for each object. We generate the regions for the perturbed environments by expanding the width and height of these regions by a factor of 1.2$\times$ for each task-relevant object. Additionally, we mandate that at least one task-relevant object (e.g., the black bowl in the \textit{put the black bowl on top of the cabinet}) exists in the expanded portion of its initial region in order to test spatial generalization of the policy. As a further test of generalization, we  introduce distractor objects into tasks. Specifically, we sample 1--2 random objects from the Libero object suite (excluding objects that are already present in the scene as well as large objects such as \textit{microwave}) and randomly place these in the scene in regions that do not collide with existing objects. We conduct evaluations under \textit{Perturbation} as well as the combination of \textit{Perturbation + Distractors}.

\subsection{Bridge Description}

\begin{figure}[t]
    \centering
    \includegraphics[width=\linewidth]{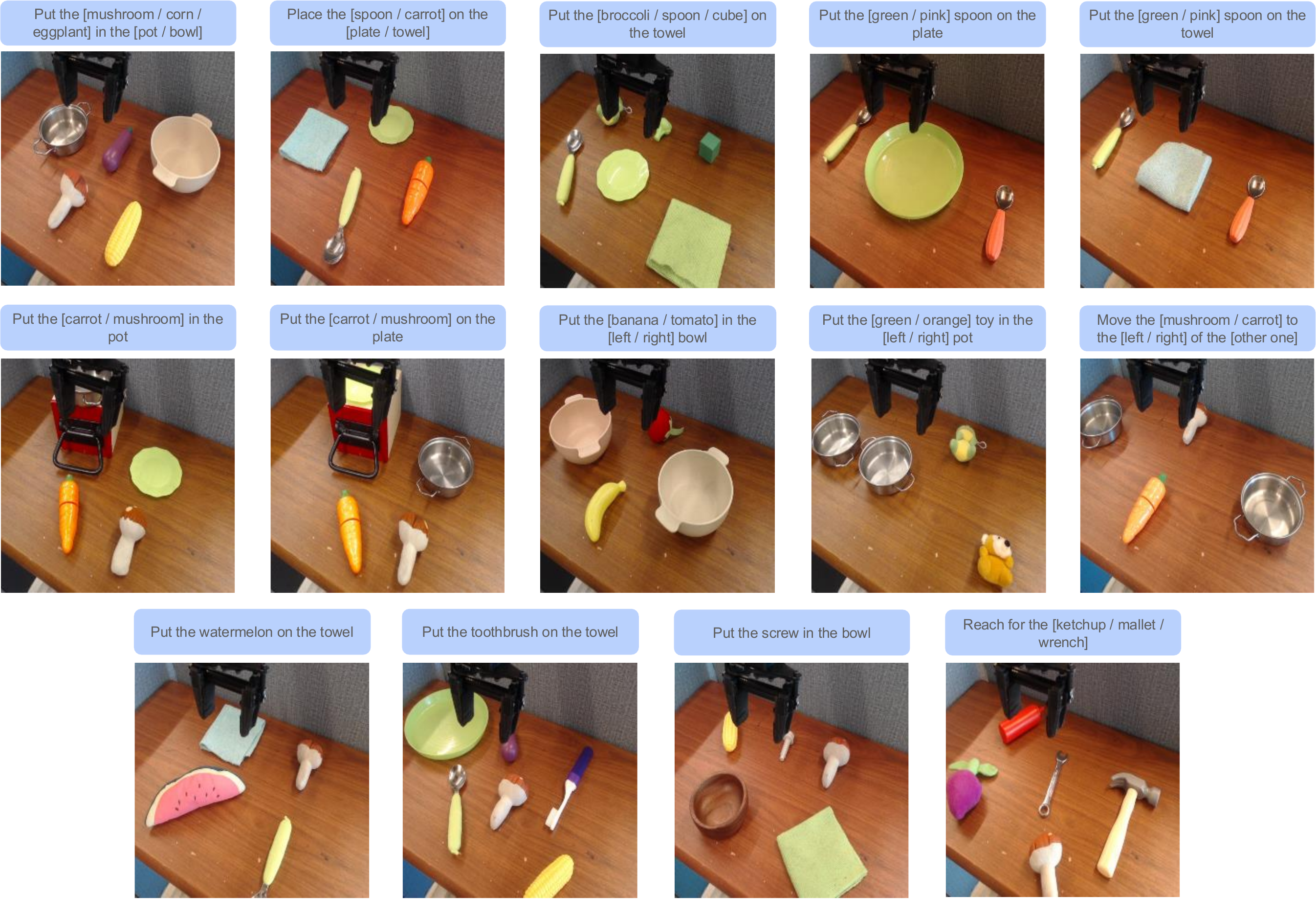}
    \caption{
        \footnotesize{Full list of task prompts used for our Bridge evaluation, as well as example starting states. We test all combinations of the words in brackets for each scene.}
    }
    \label{fig:bridge-tasks}
\end{figure}

Our real-world robot evaluations are done on a BridgeData WidowX station \citep{Ebert21-bridgeDataV1, Walke23-bridgeDataV2, EmbodimentCollaboration24-oxe}, which is a popular testbed for language-conditioned policies \citep{Kim24-openVLA, OMT23-octo}, including the original ECoT \citep{Zawalski24-ecot}. The observation space contains RGB images from a fixed, third-person camera, while the action space is a 7D vector (six elements containing delta Cartesian position and rotation, one for gripper open/close). We list all tasks in our Bridge evaluations in \cref{fig:bridge-tasks}. All permutations of the bracketed words are tested for a particular scene (and, indeed, all scenes have more than one possible task in them, making them better tests of language conditioning \citep{Gao25-starGen-taxonomy}).

\section{Embodied Reasoning Datasets}
\label{sec:app:reasoning_data}

% \textbf{Embodied reasoning annotations.}
We train all embodied reasoning policies using the same set of reasoning annotations: following prior work~\citep{Zawalski24-ecot}, we train policies to predict natural language subtasks, natural language subtask explanations, object bounding boxes, the end-effector location, and primitive movements (``move left'', ``move up''). While we stick to this particular choice of reasoning steps in this work, since it has been shown to work well in prior work~\citep{Zawalski24-ecot}, we emphasize that the choice of reasoning steps is \emph{not} a contribution of our work, and we expect the recipes we develop to transfer to other reasoning choices that have been explored in the literature \citep{Hwang24-emma, Zhao25-cotvla-visualchainofthoughtreasoning, Niu24-llarva}. For BridgeData~V2 we directly use the embodied reasoning annotations released by~\citet{Zawalski24-ecot}. For Libero, we generate reasoning annotations for the aforementioned steps using the simulator state (for bounding boxes, gripper locations, and movements), and a LLM-based synthetic data generation pipeline akin to that of~\citet{Zawalski24-ecot} (for plans, subtasks, and subtask or motion reasonings).

% KP: explain what reasoning steps we use in detail
% KP: then explain for both datasets how we generate the reasoning annotations (I guess for Bridge we can just say that we use the released annotations)

\subsection{LIBERO-90 Embodied Chain-of-Thought Pipeline and Dataset Details}
\label{app_subsec:libero-ecot}

We now describe how we reproduced the ECoT data and policy pipeline for LIBERO-90.

As with the original Bridge pipeline, we extract the following reasoning features: a list of high-level subtasks for accomplishing the overall task (\textbf{PLAN}), a natural language rationale for which subtask it should currently follow (\textbf{SUBTASK REASON, SUBTASK}, an equivalent rationale for what low-level language motion (move left, close gripper, etc) to execute (\textbf{MOVE REASON, MOVE}), object bounding boxes (\textbf{OBJECTS}), and gripper pixel coordinates (\textbf{GRIPPER}). 

These are roughly what were used in the original ECoT work, with some small changes to how they were extracted to better suite LIBERO. First, we exclude the TASK step used by \citet{Zawalski24-ecot}; as there are significantly fewer tasks in LIBERO-90 than Bridge and the evaluation instructions are always ones that are seen during training, we expect that this step is superfluous. Next, to extract object bounding boxes and gripper positions, we use the ground truth segmentation masks and object names provided by the simulator (bounding boxes are simply the maximum and minimum height and width coordinates of each mask, while the gripper position is the center of its bounding box). 
% We likewise use a slightly different programmatic approach for extracting robot language motions, adopted from \wci{TODO: @Suneel -- did you use the action labeling pipeline in your other robot CoT paper? If so, let's cite that here?}. 
Finally, we use Molmo and Llama2 to come up with and produce subtasks, subtask reasons, and move reasons \citep{Deitke24-molmo-pixmo, Touvron23-llama2}. We note these pipeline differences from the original one by \citet{Zawalski24-ecot} result in the stylistic disparities between LIBERO and Bridge reasonings (e.g., Molmo's synthetic reasonings are much ore verbose). See \cref{fig:plan-prompt}, \cref{fig:breakdown-prompt}, \cref{fig:subtask-reasoning-prompt}, and \cref{fig:motion-reasoning-prompt} for the prompts used in our LIBERO reasoning data generation pipeline. We run this pipeline on all 3,917 trajectories in our demonstration dataset, generating labels for roughly 90\% of them.

\section{Policy Architectures and Training Hyperparameters}
\label{sec:app:policy_details}

% For each environment, we use state-of-the-art VLA architectures from prior work: for Libero we use MiniVLA~\citep{Belkhale24-miniVLA}, a 1B~parameter VLA architecture based on Qwen-0.5B~\citep{Qwen24-qwen25}, that uses residual vector quantization to map from sequences of continuous actions to discrete action tokens~\citep{Belkhale24-miniVLA}; for Bridge, we follow~\citet{Zawalski24-ecot} and use OpenVLA~\citep{Kim24-openVLA}, a 7B~parameter VLA built on the Prismatic~VLM~\citep{Karamcheti24-prismatic}, which uses simple binning tokenization for the continuous action inputs. We emphasize that our policy architecture choices are merely trying to mirror prior state-of-the-art policies on the respective environments as closely as possible for the most representative results, but the training recipes we develop are \emph{agnostic} to the choice of the underlying VLA architecture and can readily be applied to embodied reasoning training on other VLA architectures.

% KP: explain in detail the VLA architectures we use in each environment and the hyperparameters we use for training

All our policy-training experiments use OpenVLA and MiniVLA \citep{Kim24-openVLA, Belkhale24-miniVLA}. The former, used for our real-robot experiments, is trained to produce discretized 7D robot manipulator actions (representing changes in Cartesian position, rotation, and gripper open/close state), expressed by binning each dimension uniformly into 256 bins, as also done by \citet{Brohan23-rt2}. The base VLM is a 7B parameter PrismaticVLM \citep{Karamcheti24-prismatic}, which uses DINOv2 and SigLIP \citep{Oquab24-dinov2,Zhai23-siglip} as visual encoders and Llama2 \citep{Touvron23-llama2} as its language backbone. The latter, used in our large-scale simulation experiments, instead discretizes an action chunk \citep{Zhao23-aloha-act} of horizon length 10 into seven tokens by passing it through a learned VQ-VAE \citep{Oord17-vq-vae} with codebook size of 256 (see \cref{app_subsec:vq-vae} for details). At inference time, the VLA likewise autoregressively generates seven action tokens, but the VQ-VAE then decodes this into a full action chunk of some specified size. MiniVLA uses a smaller base VLM that swaps out the Llama2 backbone with Qwen2.5, making the full VLA 1B parameters \citep{Qwen24-qwen25}.

Other variants of the core VLA recipe exist, such as generating actions with a diffusion head or with parallelized action embedding tokens \citep{Black24-pi0, Kim25-openVLA-oft}. In principle, these can also be used in reasoning VLAs, but we opt to use the discretization schemes used by \citet{Kim24-openVLA}, \citet{Brohan23-rt2}, and \citet{Belkhale24-miniVLA} for simplicity.

\subsection{VQ-VAE Action Chunking}
\label{app_subsec:vq-vae}
The MiniVLA codebase provides support for training policies with VQ-VAE \citep{Oord17-vq-vae} tokenization of actions. We use that code to train the tokenizer to convert ten 7D actions (so 70 elements total) into seven discrete tokens, drawn from a codebook of size 256. These tokens are then decoded into a reconstruction of the original action chunk, so the VQ-VAE effectively provides a compressed, discretized representation of the action sequence.

These seven tokens are what our MiniVLA policies predict. At inference time, after generating these tokens, they are passed through the decoder of the VQ-VAE to convert them back to a 10-step action chunk. We note that such action chunking was shown by \citet{Belkhale24-miniVLA} in the original MiniVLA LIBERO comparisons to be more effective than standard one-step action generation. However, they only consider it as a representation learning approach, as they run the action chunk ``closed loop'' (generate multiple actions, execute the first one, then generate again). However, we find that executing all ten such actions before re-querying the VLA to work better in all cases (while reducing the number of VLA queries by ten times!), so all LIBERO experiments are run as such. 

\subsection{Training Hyperparameters}

Following \citet{Zawalski24-ecot}, we mainly use the default hyperparameters for MiniVLA and OpenVLA when training reasoning policies using their respective codebases \citep{Belkhale24-miniVLA, Kim24-openVLA}. The LIBERO MiniVLA policies are trained for 200k gradient steps with batch size 128, which is approximately how long it takes for the VQ-VAE action chunking policy to reach 85\% teacher-forced action token accuracy. Some policies we tested were trained with 100k steps and achieved near-identical performance (there was no observed degradation from overtraining), but for fairness, all reported LIBERO policies use 200k steps. The Bridge OpenVLA policies are trained for 80k gradient steps with batch size 256\footnote{Note the higher batch size. We suspect that MiniVLA's Qwen backbone may have an inefficient sharding implementation, as we were unable to fit 256 samples per batch, despite said LLM being smaller.}, approximately to 95\% action token accuracy\footnote{This is higher than the MiniVLA threshold since it does NOT do action chunking, so it is easier to achieve high accuracy.}. All policies were trained on 8x A100s, except the co-trained LIBERO policy, which was trained on 8x H200s (the higher GPU memory is needed to fit the doubled batch size, as described in \cref{app_subsec:cotraining}). We used the default hyperparameters provided by the MiniVLA codebase for training the horizon length 10 VQ-VAE action chunk tokenizer, using the final checkpoint (after 200 epochs).

\section{Experiment and \Acronym{} Policy Training Details}

\subsection{Reasoning Pre-training}
\label{app_subsec:pretraining}
In reasoning pre-training, the policy takes in the observation and task language and maps it to the corresponding embodied reasoning, trained using the standard VLM NLL objective. Implementation-wise, this simply involves removing the action tokens from the end of each datapoint in the ECoT dataloader. See left part of (b) in \cref{fig:policy_variants}. In action fine-tuning, the checkpoint from the end of pre-training is loaded and tuned with the standard VLA objective. See right part of (b) in \cref{fig:policy_variants}. Note that, by doing these two learning objectives in sequence, we cannot train for the same number of total gradient steps as other approaches while also keeping the number of steps for reasonings and actions individually identical. In LIBERO, we do 100k steps of reasoning pre-training, then 200k steps of action-tuning. In Bridge, we do 20k steps of reasoning and 80k of actions.

\subsection{Reasoning Co-training}
\label{app_subsec:cotraining}
In reasoning co-training, the policy learns to map images and task language to reasonings and actions simultaneously (albeit in different training data points, i.e., actions cannot attend to reasonings, nor vice-versa). These are the same two objectives as used in reasoning pre-training and action fine-tuning, just at the same time -- each batch sample has an equal probability of being for action and reasoning prediction. In this setting, we train for 200k steps (as done for the all other LIBERO policies, minus reasoning pre-training), but double the batch size to 256. That way, the policy is trained on the same amount of reasoning and action data as our ECoT policy (wherein each batch sample contains both reasonings and actions, rather than one or the other).

\subsection{Reasoning Dropout}
\label{app_subsec:dropout}
In this case, for each batch sample with a reasoning annotation, we randomly and uniformly choose a number from zero to the total number of reasoning steps and drop out that many steps. The remaining reasoning steps (if any) are then put together into the reasoning, and then prepended to the action. Losses are assigned to both the action tokens and reasonings (if present). Thus, some fraction of training points lack reasonings, meaning that the policy is trained to sometimes map observations and tasks directly to actions (without reasonings), allowing it to behave like a standard VLA, simply by starting its generations at the start of action indicator (which, in our case, is just the string ``ACTION:'', expressed as tokens). Conversely, this also means that it can act as an ECoT reasoning policy as well by starting its generations from the start of reasoning indicator.

Other than adding train-time dropout of the reasonings, the model's inputs, outputs, and hyperparameters are identical to ECoT -- note that the illustrations in \cref{fig:policy_variants} for ECoT in (a) and reasoning dropout in (c) are identical, with the exception that the reasonings are randomly dropped when training and disabled during testing.

Notably, we find that a significant portion of Bridge transitions lack reasonings in the annotation dataset released by \citet{Zawalski24-ecot}. Thus, we find that their released policy can actually be run with reasonings dropped out at test-time. This is therefore the policy we use in the Bridge comparison.

\subsection{Reasoning Scaffolding}
\label{app_subsec:scaffolding}

This policy is trained the same way as reasoning dropout, except that no loss is assigned to reasoning tokens (that is, it does not learn to generate the reasonings, only accept them as part of the context). Critically, it also uses the titular random reasoning dropout, so the policy is able to map images (without in-context reasonings) to actions. All hyperparameters are thus the same as dropout. Note how (c) and (d) in \cref{fig:policy_variants} are identical, except the reasonings are listed as an input in the latter.

\subsection{Thinking Token Policies}
\label{app_subsec:thinking-tokens}

We now provide details for our ``thinking token'' approach, inspired by \citet{Pfau24-letsthinkdotdot} and \citet{Goyal24-thinkBeforeYouSpeak}. 

\subsubsection{Theory of Thinking Tokens}
This approach processes some number of filler thinking tokens before producing an answer, thereby increasing the expressivity by the Transformer (as all those tokens now can support additional computations). Critically though, this does \textit{not} improve the complexity class of the Transformer, which is known to be TC$^0$; it only increases the expressivity of the fixed-size network within that class \citep{Merrill23-parallelismtradeoff, Pfau24-letsthinkdotdot}. This is in contrast to true chain-of-thought reasoning, wherein a Transformer autoregressively generates different tokens. As discussed by \citet{Feng23-revealingmysteryCoT}, in that case, by generating a suitably long and appropriate sequence of tokens, Transformers can express computations in class NC$^1$, whereas such a computation cannot be solved by querying the Transformer to generate the answer directly (without intermediate computations)\footnote{Assuming TC$^0$ cannot express NC$^1$ \citep{Yao89-circuitsLocalComputation}.}.

Intuitively, being able to autoregressively generate a reasoning consisting of appropriate different tokens allows Transformers to ``lock in'' intermediate computations for all subsequent ones to attend to. \citet{Feng23-revealingmysteryCoT} discusses the case of integer arithmetic, wherein the correct intermediate steps are to encode long addition (adding corresponding digits of the numbers from right to left, carrying overflows to the next digit) within the CoT. 

In contrast, uniform (or non-uniform but low-complexity) sequences of thinking tokens do NOT enjoy this benefit. Since each thinking token is uniform, they do not ``lock in'' the computations of previous tokens; their initial embeddings do not depend on any past computations, whereas in CoT, since that token was generated and fed back into the Transformer autoregressively, this is not the case.

Notably, it is unclear to what extent this argument holds for the case of robot reasoning, or other problems wherein training the CoT to exactly represent the sequential computations needed to solve a problem is intractable. Thus, while we control for number of tokens to induce similar expressivity in our thinking token and ECoT policies, we note that they are not perfectly identical.

\subsubsection{Practical Implementation Details of Thinking Token Policies}
To train a thinking token policy, we replace the reasonings in ECoT with a similar number of thinking tokens. Following the practice of using rarely-used tokens to symbolize model actions in VLAs \citep{Kim24-openVLA}, we choose the `` .'' (space followed by a period) token as our dedicated thinking token, as it never appears in our training data and does not intuitively have any semantic meaning related to our robot tasks. This is a design choice -- we can alternatively add a brand-new extra thinking token, as done by \citet{Goyal24-thinkBeforeYouSpeak}.

Implementation wise, for each datapoint, we uniformly sample a number from 50-350, then include that many of the thinking tokens as part of the prompt (i.e., no loss is assigned to them). This is therefore functionally identical to the standard VLA training process, just with the input prompts altered with appended thinking tokens -- note the similarity between the standard VLA in (a) and the thinking token policy in (e) in \cref{fig:policy_variants}. Note that the VLAs we use have decoder-only LLM backbones, meaning prompt and generated tokens are treated identically (only differing in which ones have losses assigned to them during training). This is not the case for some VLA architectures, like those based on PaliGemma \citep{Beyer24-paligemma, Steiner24-paligemma2}, which applies non-causal attention to prompt tokens only.

At test time, we append varying number of thinking tokens to the prompt, followed by the start of action prefix. The policy therefore only generates the seven action tokens, but gets to attend to the representations of both the policy inputs and all intermediate thinking tokens (thereby expanding its expressivity).

\begin{figure}[t]
    {\tiny\texttt{\input{text_figs/PlanPrompt.txt}}}
    \caption{
        \footnotesize{The Llama2 prompt for generating a plan (list of subtasks) from an instruction.}
    }
    \label{fig:plan-prompt}
\end{figure}

\begin{figure}[t]
    {\tiny\texttt{\input{text_figs/BreakdownPrompt.txt}}}
    \caption{
        \footnotesize{The Llama2 prompt for aligning the generated subtasks to specific time steps.}
    }
    \label{fig:breakdown-prompt}
\end{figure}

\begin{figure}[t]
    {\tiny\texttt{\input{text_figs/SubtaskReasoningPrompt.txt}}}
    \caption{
        \footnotesize{The Molmo prompt for generating subtask reasonings for each subtask generated via \cref{fig:plan-prompt}. The VLM also accepts the current observation image.}
    }
    \label{fig:subtask-reasoning-prompt}
\end{figure}

\begin{figure}[t]
    {\tiny\texttt{\input{text_figs/MotionReasoningPrompt.txt}}}
    \caption{
        \footnotesize{The Molmo prompt for generating movement reasonings. The VLM also accepts the current observation image.}
    }
    \label{fig:motion-reasoning-prompt}
\end{figure}

%% file: text_figs/LiberoTasks1.txt
close the top drawer of the cabinet

close the top drawer of the cabinet and put the black bowl on top of it

put the black bowl in the top drawer of the cabinet

put the butter at the back in the top drawer of the cabinet and close it

put the butter at the front in the top drawer of the cabinet and close it

put the chocolate pudding in the top drawer of the cabinet and close it

open the bottom drawer of the cabinet

open the top drawer of the cabinet

open the top drawer of the cabinet and put the bowl in it

put the black bowl on the plate

put the black bowl on top of the cabinet

open the top drawer of the cabinet

put the black bowl at the back on the plate

put the black bowl at the front on the plate

put the middle black bowl on the plate

put the middle black bowl on top of the cabinet

stack the black bowl at the front on the black bowl in the middle

stack the middle black bowl on the back black bowl

put the frying pan on the stove

put the moka pot on the stove

turn on the stove

turn on the stove and put the frying pan on it

close the bottom drawer of the cabinet

close the bottom drawer of the cabinet and open the top drawer

put the black bowl in the bottom drawer of the cabinet

put the black bowl on top of the cabinet

put the wine bottle in the bottom drawer of the cabinet

put the wine bottle on the wine rack

close the top drawer of the cabinet

put the black bowl in the top drawer of the cabinet

put the black bowl on the plate

put the black bowl on top of the cabinet

put the ketchup in the top drawer of the cabinet

close the microwave

put the yellow and white mug to the front of the white mug

open the microwave

put the white bowl on the plate

put the white bowl to the right of the plate

put the right moka pot on the stove

turn off the stove

put the frying pan on the cabinet shelf

put the frying pan on top of the cabinet

put the frying pan under the cabinet shelf

put the white bowl on top of the cabinet

turn on the stove

%% file: text_figs/LiberoTasks2.txt
turn on the stove and put the frying pan on it

pick up the alphabet soup and put it in the basket

pick up the cream cheese box and put it in the basket

pick up the ketchup and put it in the basket

pick up the tomato sauce and put it in the basket

pick up the alphabet soup and put it in the basket

pick up the butter and put it in the basket

pick up the milk and put it in the basket

pick up the orange juice and put it in the basket

pick up the tomato sauce and put it in the basket

pick up the alphabet soup and put it in the tray

pick up the butter and put it in the tray

pick up the cream cheese and put it in the tray

pick up the ketchup and put it in the tray

pick up the tomato sauce and put it in the tray

pick up the black bowl on the left and put it in the tray

pick up the chocolate pudding and put it in the tray

pick up the salad dressing and put it in the tray

stack the left bowl on the right bowl and place them in the tray

stack the right bowl on the left bowl and place them in the tray

put the red mug on the left plate

put the red mug on the right plate

put the white mug on the left plate

put the yellow and white mug on the right plate

put the chocolate pudding to the left of the plate

put the chocolate pudding to the right of the plate

put the red mug on the plate

put the white mug on the plate

pick up the book and place it in the front compartment of the caddy

pick up the book and place it in the left compartment of the caddy

pick up the book and place it in the right compartment of the caddy

pick up the yellow and white mug and place it to the right of the caddy

pick up the book and place it in the back compartment of the caddy

pick up the book and place it in the front compartment of the caddy

pick up the book and place it in the left compartment of the caddy

pick up the book and place it in the right compartment of the caddy

pick up the book and place it in the front compartment of the caddy

pick up the book and place it in the left compartment of the caddy

pick up the book and place it in the right compartment of the caddy

pick up the red mug and place it to the right of the caddy

pick up the white mug and place it to the right of the caddy

pick up the book in the middle and place it on the cabinet shelf

pick up the book on the left and place it on top of the shelf

pick up the book on the right and place it on the cabinet shelf

pick up the book on the right and place it under the cabinet shelf

%% file: text_figs/PlanPrompt.txt
I am trying to give a robotic arm a short list of high-level steps to complete the task: '\{instruction\}.' 
The robotic arm is in a scene containing the following \{number of objects\} objects: 

\{list of object names\}

Break down this instruction into a Python list of high-level steps. 
These steps should describe the broad physical actions the robot should take to accomplish the task, not any perceptual steps. 
For example, for the task 'Place the mushroom in the pot,' a possible list of tasks could be: ['move to the mushroom', 'grasp the mushroom', 'move the mushroom to the pot', 'release the mushroom into the pot']

%% file: text_figs/BreakdownPrompt.txt
I want to analyze each step my robot took to accomplish the task: '\{instruction\}.' To accomplish this task, it takes these subtasks in this order:

\{numbered list of subtasks\}

Here is a mapping from timestep to the motion that my robot took:

\{list of time step indices to language motions, deduplicated\}

Please provide me with a Python dictionary mapping from each of the \{number of subtasks\} subtasks (as strings) to the integer timestep when that subtask begins. Remember that subtasks are in order, so later subtasks must map to later timesteps. The first subtask should always start at step 0.

%% file: text_figs/SubtaskReasoningPrompt.txt
The robot's task is: '\{instruction\}.' The robot's plan for completing this instruction is:

\{numbered list of subtasks\}

Based on the position of objects, the robot's pose relative to those objects, the layout of the scene, explain why the robot's current subtask is: \{current subtask\}. Answer concisely in a single sentence.

%% file: text_figs/MotionReasoningPrompt.txt
I want you to explain a robot's actions to me. The robot's task is: '\{instruction\}.' The robot's plan for completing this instruction is: 

\{numbered list of subtasks\}

The robot's current subtask in achieving that task is: \{subtask\}. 

The robot just took the following \{number of past deduped language motion actions\} actions: 

\{numbered list of language motion actions\}

Based on the position of objects, the robot's pose relative to those objects, the layout of the scene, why should the robot '\{current language motion\}' in order to complete that subtask in this situation?

Answer with something like: '<some reason>, thus the robot should <some motion>'. Answer specifically, but concisely in one sentence.